\documentclass{article}

    \usepackage[preprint]{neurips_2025}

\usepackage[utf8]{inputenc} 
\usepackage[T1]{fontenc}    
\usepackage{hyperref}       
\usepackage{url}            
\usepackage{booktabs}       
\usepackage{amsfonts}       
\usepackage{nicefrac}       
\usepackage{microtype}      
\usepackage{xcolor}         

\title{AGI-Elo: How Far Are We From Mastering A Task? }

\author{
    Shuo Sun$^{1,3}$ \quad
    Yimin Zhao$^{1}$ \quad
    Christina Dao Wen Lee$^{1}$ \quad
    Jiawei Sun$^{1}$ \quad
    Chengran Yuan$^{1}$ \\
    \textbf{Zefan Huang$^{1,3}$} \quad
    \textbf{Dongen Li$^{1,3}$} \quad
    \textbf{Justin KW Yeoh$^{1}$} \quad
    \textbf{Alok Prakash$^{3}$} \\
    \textbf{Thomas W. Malone$^{2,3}$} \quad
    \textbf{Marcelo H. Ang Jr.$^{1,3}$} \\
    \\
    $^1$National University of Singapore \quad
    $^2$Massachusetts Institute of Technology \\
    $^3$Singapore MIT Alliance for Research and Technology \\
    \texttt{\{shuo.sun,yimin.zhao,christinaldw,sunjiawei,chengran.yuan,} \\ \texttt{huangzefan,li.dongen\}@u.nus.edu} \quad
    \texttt{alok.prakash@smart.mit.edu} \\
    \texttt{malone@mit.edu} \quad
    \texttt{\{justinyeoh,mpeangh\}@nus.edu.sg}
}

\usepackage{graphicx}
\usepackage{tabularx}
\usepackage{multirow}
\usepackage{tabulary}
\usepackage{subfig}
\usepackage{amsmath}

\begin{document}

\maketitle

\begin{abstract}
As the field progresses toward Artificial General Intelligence (AGI), there is a pressing need for more comprehensive and insightful evaluation frameworks that go beyond aggregate performance metrics. 
This paper introduces a unified rating system that jointly models the difficulty of individual test cases and the competency of AI models (or humans) across vision, language, and action domains. 
Unlike existing metrics that focus solely on models, our approach allows for fine-grained, difficulty-aware evaluations through competitive interactions between models and tasks, capturing both the long-tail distribution of real-world challenges and the competency gap between current models and full task mastery. 
We validate the generalizability and robustness of our system through extensive experiments on multiple established datasets and models across distinct AGI domains. 
The resulting rating distributions offer novel perspectives and interpretable insights into task difficulty, model progression, and the outstanding challenges that remain on the path to achieving full AGI task mastery. We have made our code and results publicly available at \url{https://ss47816.github.io/AGI-Elo/}.
\end{abstract}

\begin{figure}[ht]
    \centering
    \includegraphics[width=0.8\textwidth]{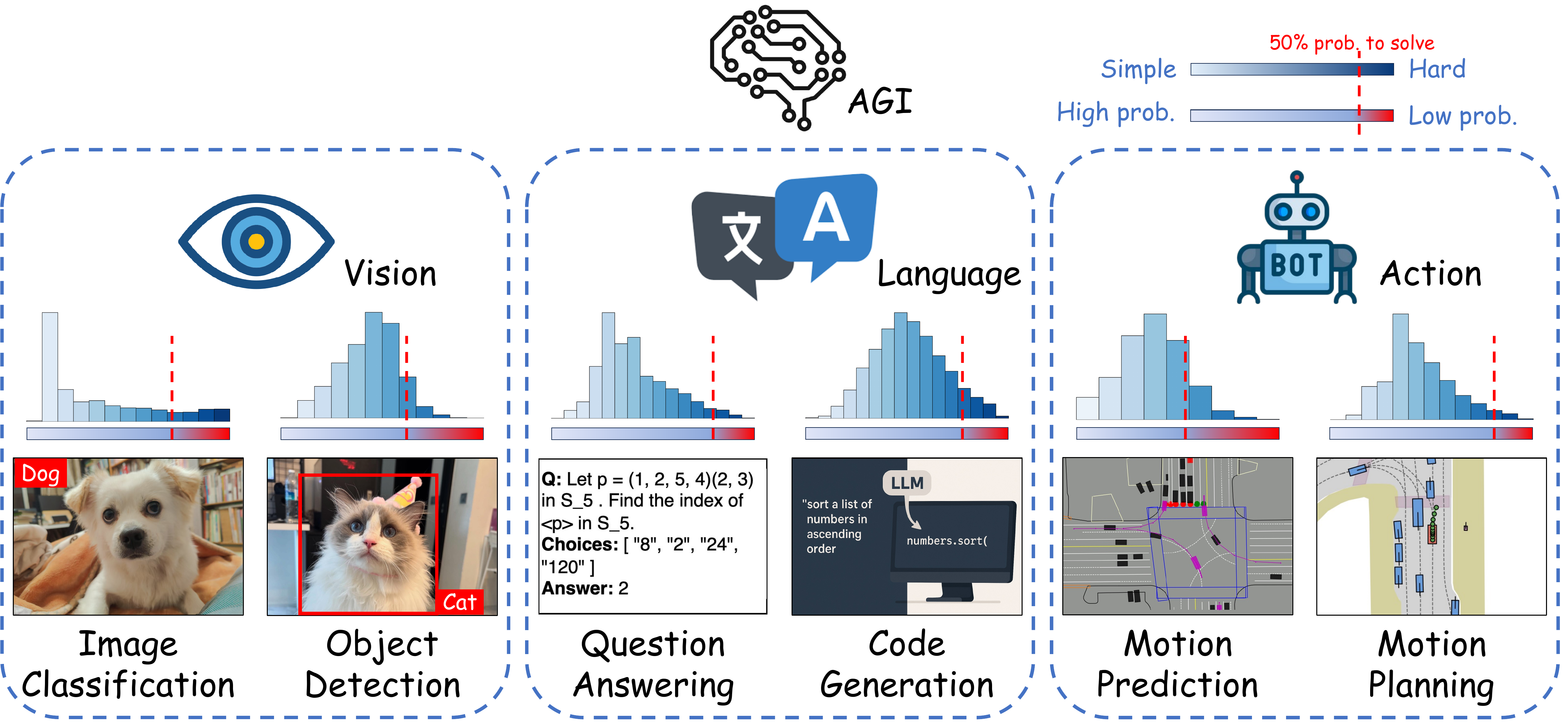}
    \caption{
        In this paper, we address long-standing questions regarding the current capabilities of AGI and humans on challenging tasks by proposing a standardized framework to quantitatively assess task difficulty, evaluate AGI competency, and identify gaps to task mastery. 
    }
    \label{fig:intro}
\end{figure}

\section{Introduction}

As Artificial General Intelligence (AGI) begins to replace traditional Artificial Intelligence (AI) in our everyday lives, there is a growing need to systematically evaluate state-of-the-art (SOTA) AI models across a diverse range of tasks. These tasks span three fundamental domains: vision, language, and action. A crucial aspect of this evaluation is understanding not only the performance of AI models but also considering the inherent difficulty of the tasks they attempt to solve, and identifying the competency gap between the current models and remaining unsolved difficult cases. As illustrated in \autoref{fig:intro}, this paper aims to quantitatively address three key questions simultaneously:
\begin{itemize}
    \item What is the difficulty of each test case within a task or a dataset?
    \item What is the competency of an AI model (or a human) on a given task? 
    \item How far are the current SOTA models from fully mastering a task? 
\end{itemize}

\subsection{Existing gaps in AI evaluation}

Despite the extensive research on AI benchmarking, several fundamental gaps remain unaddressed:

\textbf{Quantifying task and test case difficulties}: Defining and measuring the difficulty of an entire task (e.g., a dataset) or an individual test case (e.g., a single image, question, or driving scenario) remains a fundamental challenge. While a range of heuristic proxies have been explored, such as curriculum learning signals~\cite{bengio2009curriculum}, input characteristics~\cite{spitkovsky2010baby, gururangan2018annotation}, training loss~\cite{han2018co, arazo2019unsupervised, shen2019learning}, model confidence~\cite{hovy2013learning}, prediction variance~\cite{chang2017active}, and information-theoretic measures~\cite{varshney2022ildae}—these methods often rely on task-specific assumptions or model-dependent signals. A unified, systematic framework for quantifying difficulty consistent across tasks and interpretable from both AI and human perspectives is still lacking.

\textbf{Difficulty-aware \& predictive metric for AI}: Most public benchmarks and datasets~\cite{mscoco, Krizhevsky09-cifar, deng2009imagenet, Geiger2012CVPR-kitti, nuscenes, nuplan, waymodataset, NAVSIM} rely on task-specific metrics such as accuracy, mean Average Precision (mAP), and success rate to evaluate model performance. However, these metrics typically capture only aggregated performance across the dataset, providing relative comparisons between models rather than predictive indicators of how well an AI model (or a human) would perform on individual test cases of varying difficulty. 
This averaging effect obscures the underlying distribution of task difficulty and limits our understanding of a model's capacity to adapt to diverse and complex scenarios.


\textbf{Progress over the long-tail in real-world tasks}: Many real-world tasks exhibit a long-tail distribution, where certain test cases are significantly more challenging than others~\cite{longtaillearning2024wang}. Identifying these difficult cases remains non-trivial, and existing benchmarks do not provide a systematic way to measure the length of the "tail", which is, how much further AI models must progress before confidently claiming task mastery at a well-defined confidence interval (e.g., 50\%, 90\%, or 99\%) . 

\subsection{Our contributions}
To address these gaps identified, we propose a rating system that jointly models task difficulty and model competency in a unified, probabilistic manner. Our key contributions are as follows:

    
    
    


\begin{enumerate}
    \item \textbf{A task-agnostic rating system for AGI evaluations}: We introduce a rating system that simultaneously models test case difficulties and model competencies using a probabilistic approach. The rating of each test case or model is modeled as a normal distribution, which is constantly updated by a series of competitive matches between models and test cases.
    
    \item \textbf{Unified measurement of test case difficulty and model competency}: Our framework provides a principled way to quantitatively estimate the difficulty of individual test cases and the comparative competency of intelligent agents (models or humans) simultaneously.
    
    \item \textbf{Extensive experiments across domains}:  Extensive experiments were conducted across the 3 AGI domains: vision, language, and action. To this end, we considered 6 well-established datasets using 7-20 models/humans that demonstrated effectiveness.
    
    \item \textbf{Comprehensive evaluation and predictive insights}: By establishing a singular rating system for each of the AGI tasks, we analyze the rating distribution of test cases and the model ratings to identify the task difficulty distributions and long-tail characteristics. With this, we can conclude the competency gap from current models to fully mastering a task.

\end{enumerate}
By establishing a robust and predictive rating system, our work provides a new perspective on AI evaluation, paving the way for a more comprehensive understanding of AI capabilities and limitations as we move toward AGI. 
\section{Rating systems explained}

\subsection{Conventional rating systems}

Rating systems are commonly used to estimate the relative skill or performance of players based on outcomes of pairwise (or multiplayer) matches. 
After each match, the ranking system awards rating points to the winning side and deducts rating points from the losing side in a \textit{zero-sum} fashion based on the match result. 

\textit{Elo}~\cite{elo1967proposed} is the foundational rating system originally developed for chess. It updates the ratings of both players based on the match score, assuming a logistic model of win probability. Given two players \( A \) and \( B \) with ratings \( R_A \) and \( R_B \), the expected score of \( A \) can be computed as
\begin{equation}
\label{eq:e}
    \mathbb{E}[S_A] = \frac{1}{1 + 10^{(R_B - R_A)/400}}.
\end{equation}
The rating update formula is given by:
\begin{equation}
    R_A \leftarrow R_A + K(S_A - \mathbb{E}[S_A]),
\end{equation}
where \( K \) is a sensitivity parameter and \( S_A \in [0, 1] \) is the match score of \( A \).

\textit{Glicko}~\cite{glickman1999glicko} extends the Elo system by modeling a player's rating as a Gaussian belief distribution characterized by a mean \( \mu \) and a Rating Deviation (RD) \( \sigma\), which quantifies the uncertainty in rating. Ratings with higher RD values are updated with a higher magnitude as compared to players with low RD, whose ratings will be more stable. 


\subsection{Properties \& utilities}

\textbf{Probabilistic prediction}: A key utility of rating systems is their predictive power. Given two players’ ratings, the system can estimate the probability of each outcome based on \autoref{eq:e}. 

\textbf{Translation-invariant}: Rating systems are translation-invariant: shifting all ratings by a constant value does not affect the expected outcome. Only the relative difference in ratings between the two players determine the result, as the absolute scale is arbitrary and does not influence ranking behavior.

\textbf{Transitivity}: A desirable property of rating systems is transitivity: if player \( A \) consistently beats \( B \), and \( B \) consistently beats \( C \), then we expect \( A \) to have a higher rating than \( C \). Transitivity enables the construction of a consistent global ranking across many players without requiring exhaustive pairwise evaluation.

\textbf{Efficient placement}: Only a small number of matches is required to determine the rating of a new player to the system. Efficient placement of new players with minimal evaluations is critical in large-scale settings. 

\section{AGI-Elo rating system design}

\begin{figure}[bhtp]
    \centering
    \includegraphics[width=0.9\textwidth]{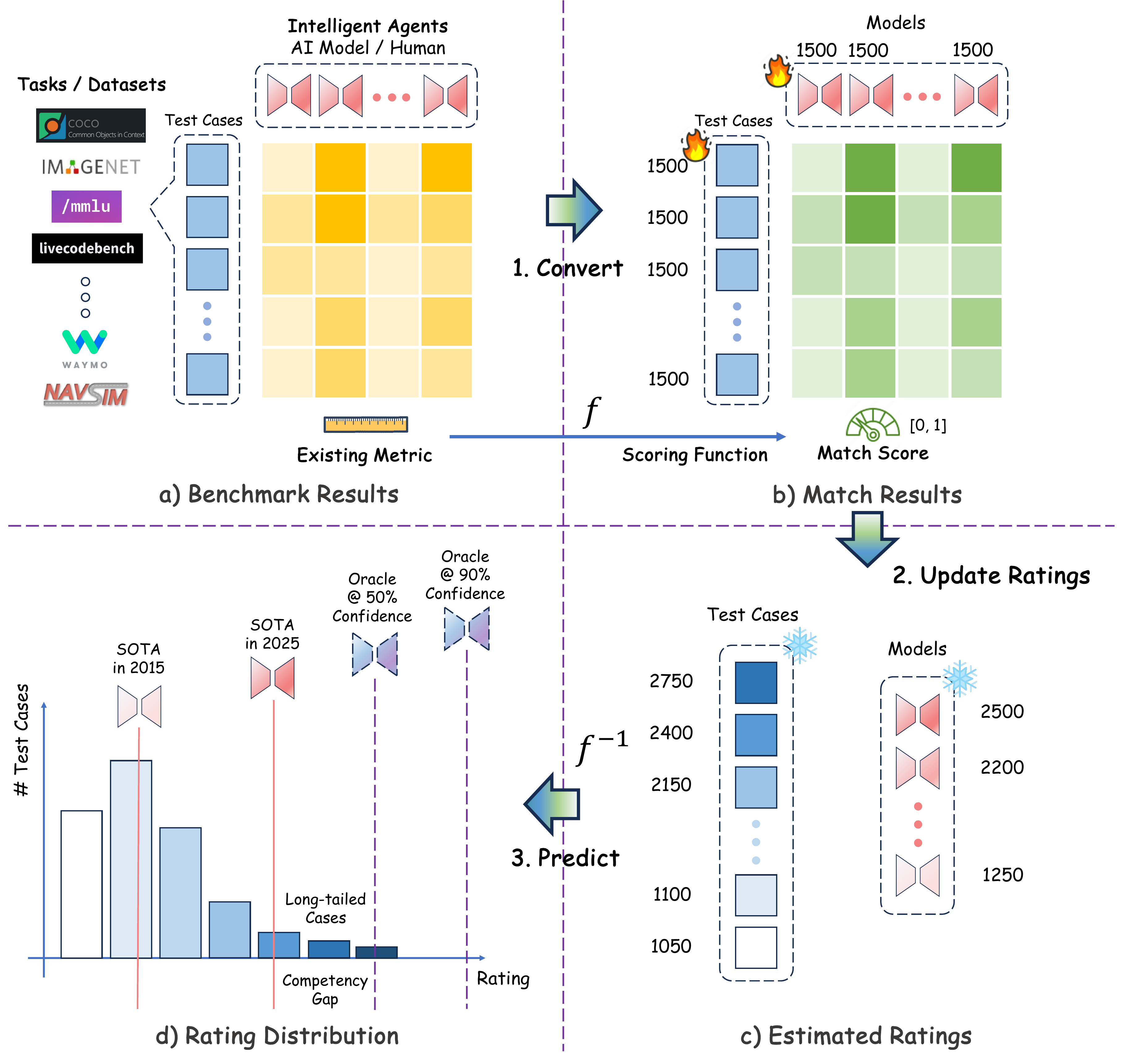}
    \caption{
        Illustration of the proposed \textit{AGI-Elo} rating system.
    }
    \label{fig:system}
\end{figure}

The proposed \textit{AGI-Elo} rating system consists of three main steps, including the conversion from benchmark results to match results, the update of models' and test cases' ratings based on match results, and the prediction of model competencies, as illustrated by the three arrows in \autoref{fig:system}. 

\subsection{Test cases vs. agents}

Conventional rating systems are primarily designed for \textit{homogeneous} agents that can freely compete against one another in direct, one-on-one matches. In chess, humans and computers are assumed to be in the same category and compete directly, sharing comparable characteristics that make such matches meaningful. 

However, our proposed rating system diverges significantly as it is designed for matches between \textit{heterogeneous} agents, in a similar fashion to Item Response Theory (IRT)~\cite{lord2008statistical}, which models the probability that an agent (human or model) with a certain ability level correctly solves a test case as:
\begin{equation}
    P(\text{correct} \mid \alpha, \beta, R_t, R_a) = \frac{1}{1 + \beta^{-\alpha(R_t - R_a)}}
\end{equation}
where \(R_t\) and \(R_a\) presents the difficulty of the test case and the ability of the agent, and \(\alpha = 1/400\), \(\beta = 10\) are assigned to follow existing conventions used in chess rating systems. 

Specifically, \textit{AGI-Elo}, defines two distinct player types: \textbf{test cases} and \textbf{agents} (i.e., models or humans), and players can only engage in inter-category matches. A test case can be matched against an agent, but never \textit{directly compete} with another test case; similarly, agents cannot compete with each other.


To enable the joint estimation of test case and agent ratings, \textit{AGI-Elo} leverages the transitivity property of rating systems, under the assumption that the transitivity property remains valid in our \textit{heterogeneous} agent setting (an assumption later supported by our experimental results in \autoref{sec:reliability}).
By observing the outcomes of inter-category matches, our rating system simultaneously infers ratings for both test cases and agents. Consequently, players within the same category are evaluated indirectly, with their relative ratings inferred through shared interactions with players from the opposing category. 

Furthermore, our system explicitly incorporates the ratings of the intermediary category during the evaluation process. In particular, the rating of a test case plays a critical role in adjusting model ratings. For example, if a model fails on an easy (i.e., low-rated) test case, it is penalized more heavily than if it fails on a difficult (i.e., high-rated) one.
By accounting for the inherent difficulty of each test case, the system avoids treating all errors equally, thereby preventing serious overestimation or underestimation of model competency in the presence of exceptionally easy or hard examples.
A key advantage of this rating system design is that model ratings are anchored to the empirical difficulty distribution of test cases. Moreover, the performance of any model on any test case can be quantitatively predicted.



\subsection{Conversion to match results}

For any given task, let \( M \in \mathbb{R} \) denote a task-specific performance metric (e.g., accuracy, mAP), and let \( f: \mathbb{R} \to [0, 1] \) be a scoring function that maps \( M \) to a normalized match score \( s \in [0, 1] \). We define:
\begin{equation}
    S = f(M)
\end{equation}

The primary objective of the function \( f \) is to transform arbitrary task-specific metrics into a unified, continuous match score space, facilitating consistent comparison across matches. Once ratings are established in this normalized space, the inverse function \( f^{-1} : [0, 1] \to \mathbb{R} \) can be used to project predicted match scores back into the original metric space, yielding an interpretable predicted performance:
\begin{equation}
    \hat{M} = f^{-1}(S) 
\end{equation}

To support generalization across diverse tasks and datasets, the scoring function \( f \) can be tailored to the specific characteristics and scale of the underlying metric \( M \). This design ensures that our approach remains broadly applicable with minimal task-specific adjustments.

\subsection{Rating update}


To determine the appropriate rating adjustment after each match, we model the rating \(R\) of each player (whether a test case or a model) as a normal distribution \(R \sim \mathcal{N}(\mu, \sigma^2)\) with a mean \(\mu\) representing its rating score and a standard deviation \(\sigma\) representing the uncertainty in our estimate, following the Glicko system~\cite{glickman1999glicko}. Initially, all models and test cases are assigned the same starting ratings. 
After each rated match, the \(\mu\) and \(\sigma\) of both players are updated based on the match outcome. 
For each opponent \(j\), the impact factor \(g(\sigma_j)\), which adjusts the weight of the match outcome based on the opponent's uncertainty, is defined as:
\begin{equation}
    g(\sigma_j) = \frac{1}{\sqrt{1 + \frac{3 q^2 \sigma_j^2}{\pi^2}}}
\end{equation}
where \(q = \frac{\ln(10)}{400} \approx 0.0057565\).
The expected outcome of player \(i\) against opponent \(j\) is:
\begin{equation}
    E_{ij} = \frac{1}{1 + 10^{-g(\sigma_j) (\mu_i - \mu_j)/400}}
\end{equation}
After a rated match where player \(i\) competes against multiple opponents \(j\), the new rating is updated as:
\begin{equation}
    \mu_i \leftarrow \mu_i + \frac{q}{\frac{1}{\sigma_i^2} + \sum_j g(\sigma_j)^2 E_{ij} (1 - E_{ij})} \sum_j g(\sigma_j) (S_{ij} - E_{ij})
\end{equation}
where \(S_{ij} \in [0,1]\) represents the actual match score.
The updated rating deviation is given by:
\begin{equation}
    \sigma_i \leftarrow \left( \frac{1}{\sigma_i^2} + \sum_j g(\sigma_j)^2 E_{ij} (1 - E_{ij}) \right)^{-1/2}
\end{equation}


After a sufficient number of matches, ideally when all models have competed against all test cases, the ratings of both models and test cases should converge to stable values that reflect their respective competency and difficulty levels.

\subsection{Prediction}

With the ratings of both models and test cases determined, we can leverage the properties of the rating system to make the following predictions:

\textbf{Agent performance}: The expected performance \(\mathbb{E}[M_a]\) of an agent \(a\) in the original metric space on a test case \(t\) can be estimated as:
\begin{equation}
    \mathbb{E}[M_a] = f^{-1}(\mathbb{E}[S_a]) = f^{-1} \left( \frac{1}{1 + 10^{(R_t - R_a)/400}} \right),
\end{equation}
where \(\mathbb{E}[S_a]\) denotes the expected match outcome of agent \(a\), and \(R_a\), \(R_t\) represent the mean rating values of the agent and the test case, respectively.

\textbf{Long-tailed test cases beyond an agent’s competency}: The set of test cases on which agent \(a\) is expected to achieve a performance below a threshold \(M_{\theta}\) (in the original metric space) is defined as:
\begin{equation}
    \mathcal{T}^{\text{hard}}_{a, M_{\theta}} = \left\{
        t \in \mathcal{T} \ \middle| \ f^{-1} \left( \frac{1}{1 + 10^{(R_t - R_a)/400}} \right) < M_{\theta} 
    \right\},
\end{equation}
where \(\mathcal{T}\) denotes the complete set of test cases in the dataset.

\textbf{Agent's competency gap to full task mastery}: 
In AI and machine learning, an \textit{oracle} typically refers to a model that achieves ideal performance or provides ground-truth answers for a given task. Assuming the dataset is a faithful miniature reflection of the real-world distribution of test cases, the oracle's performance on the most difficult test case in the dataset serves as a proxy for its worst-case performance in the real world.
In the context of this paper, we further quantify an oracle using either a performance threshold \(S_{\theta}\) in the match score space or a corresponding threshold \(M_{\theta}\) in the original metric space. 
For example, a hypothetical \textit{oracle @ \(M_{\theta}\) mastery} is defined as a model capable of achieving at least \(M_{\theta}\) performance, or equivalently, at least \(S_{\theta} \times 100\%\) confidence in solving, all test cases in the task. The rating required for such an oracle can be estimated as:
\begin{equation}
\begin{aligned}
    R_{\textit{oracle@}S_{\theta}} &\geq R_{t, \text{max}} - 400 \cdot \log_{10} \left( \frac{1 - S_{\theta}}{S_{\theta}} \right), \\
\end{aligned}
\end{equation}
where \(R_{t, \text{max}} = \max\{R_t \mid t \in \mathcal{T}\}\) denotes the rating of the hardest test case in the dataset.

The competency gap for an agent \(a\) to reach this oracle-level performance is defined as:
\begin{equation}
    \text{Competency Gap} = R_{\textit{oracle@}S_{\theta}} - R_a,
\end{equation}
which quantifies how much the agent’s rating must improve in order to achieve the desired level of task mastery.

    

\section{Experiments}

\subsection{Experimental setup}
We selected six representative tasks spanning three core AGI domains—vision, language, and action: image classification, object detection, question answering, code generation, motion prediction, and motion planning. For each task, we chose the most widely adopted dataset: ImageNet~\cite{deng2009imagenet}, COCO~\cite{mscoco}, MMLU~\cite{MMLU}, LiveCodeBench~\cite{jain2024livecodebench}, Waymo~\cite{waymodataset}, and NAVSIM~\cite{NAVSIM}, respectively.

The specific agents evaluated, as well as the evaluation metrics and scoring functions used for each dataset, are detailed in \autoref{appendix:b}. Notably, the motion planning task includes a \textit{human expert} as one of the evaluated agents, alongside AI models. 
All players (both agents and test cases) are initialized with a rating of \(R \sim \mathcal{N}(1500, 350^2)\). During the rating update step, the order of matches is fully randomized to ensure smooth and unbiased convergence of ratings.

\subsection{Rating distributions}

\begin{figure}[ht]
    \centering
    \subfloat[Image classification: ImageNet~\cite{deng2009imagenet}]{\includegraphics[width=0.5\textwidth]{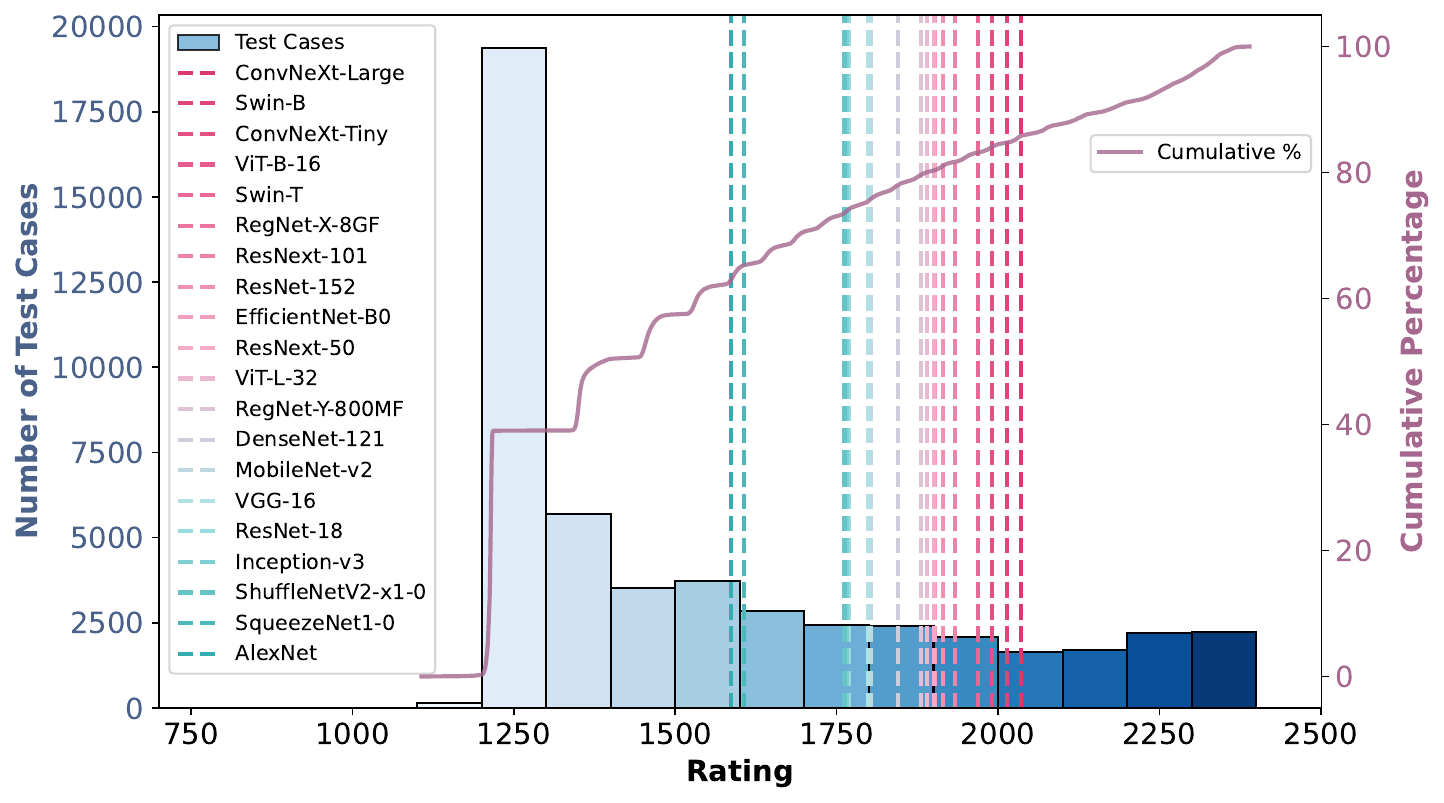}}
    \hfill
    \subfloat[Object detection: COCO~\cite{mscoco}]{\includegraphics[width=0.5\textwidth]{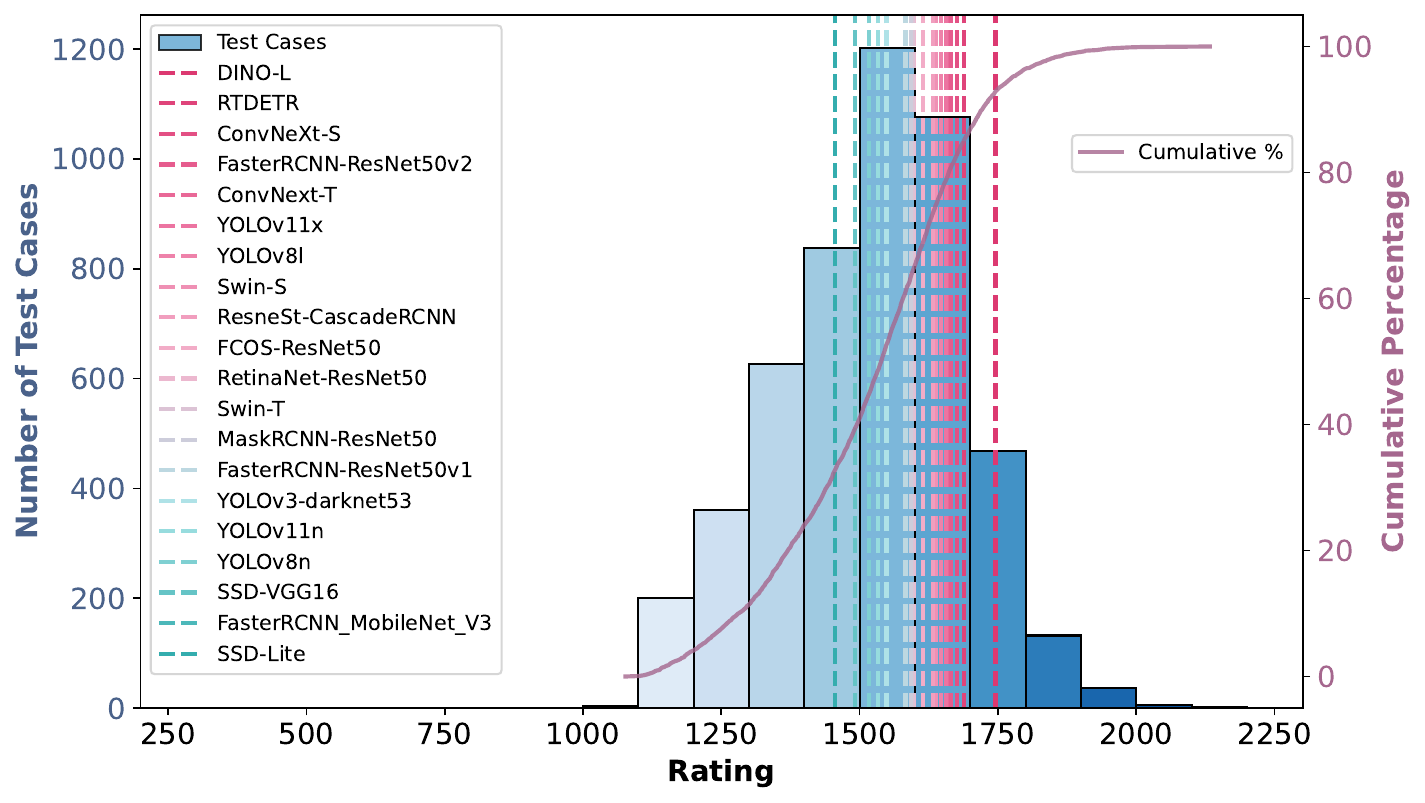}}
    \vspace{0.1pt}
    \subfloat[Question answering: MMLU~\cite{MMLU}]{\includegraphics[width=0.5\textwidth]{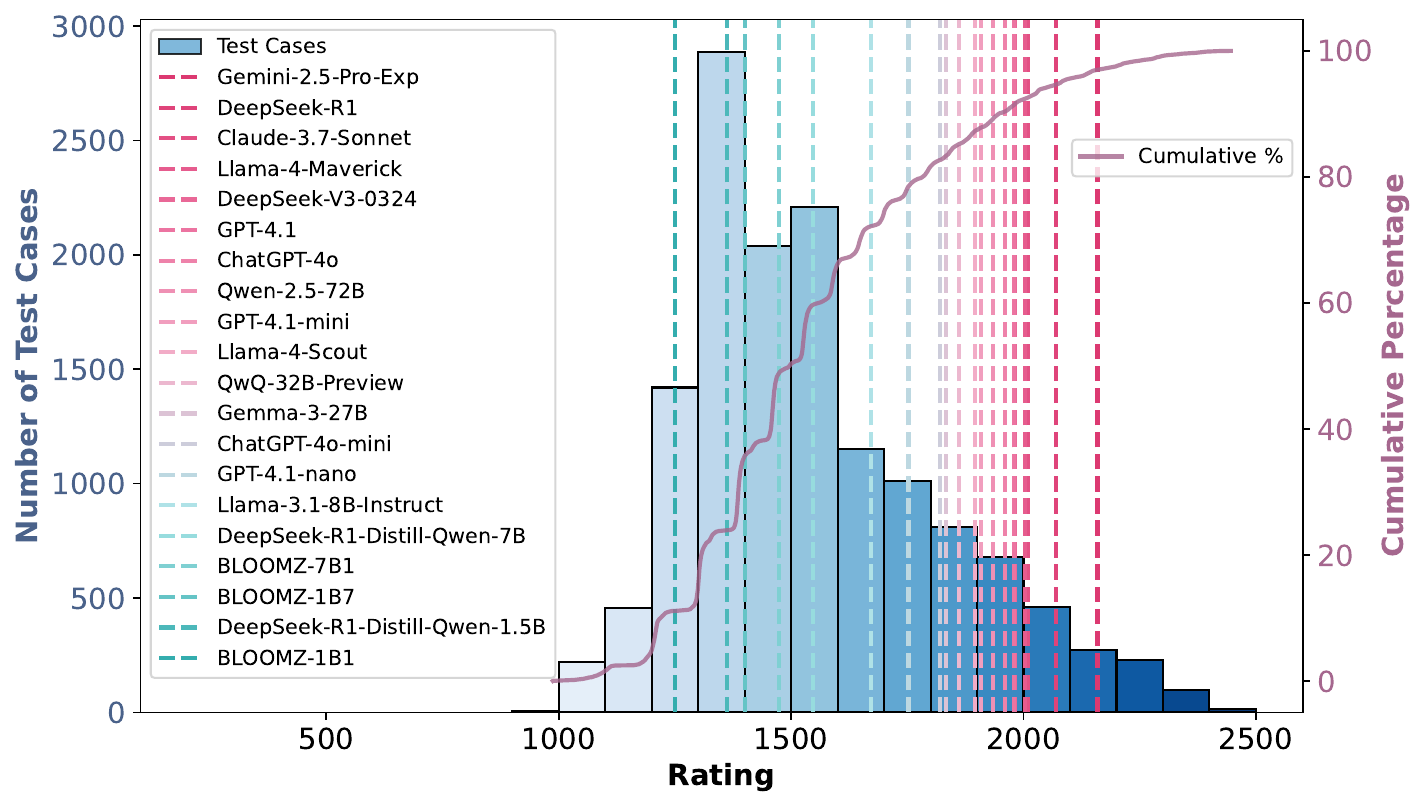}}
    \hfill
    \subfloat[Code generation: LiveCodeBench~\cite{jain2024livecodebench}]{\includegraphics[width=0.5\textwidth]{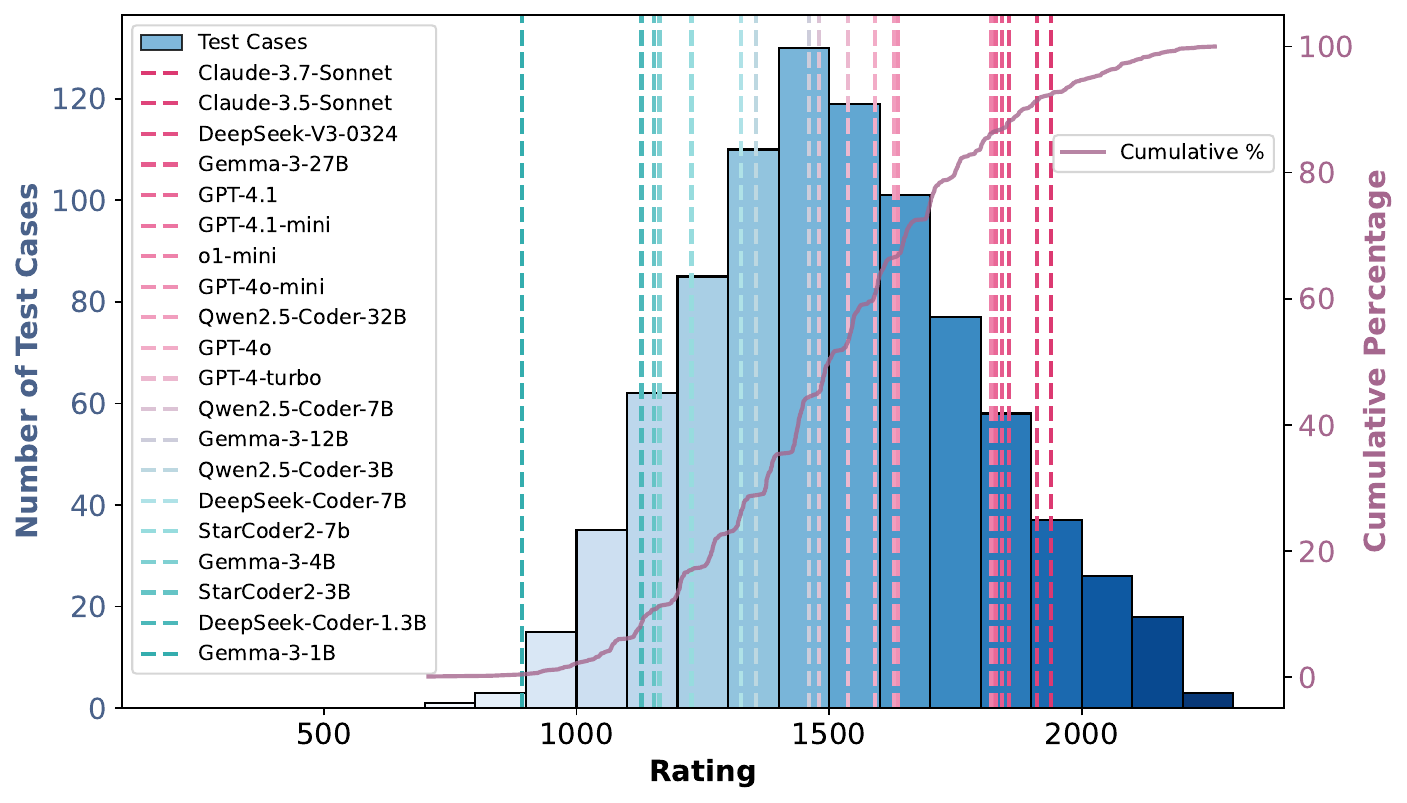}}
    \vspace{0.1pt}
    \subfloat[Motion prediction: Waymo~\cite{waymodataset}]{\includegraphics[width=0.5\textwidth]{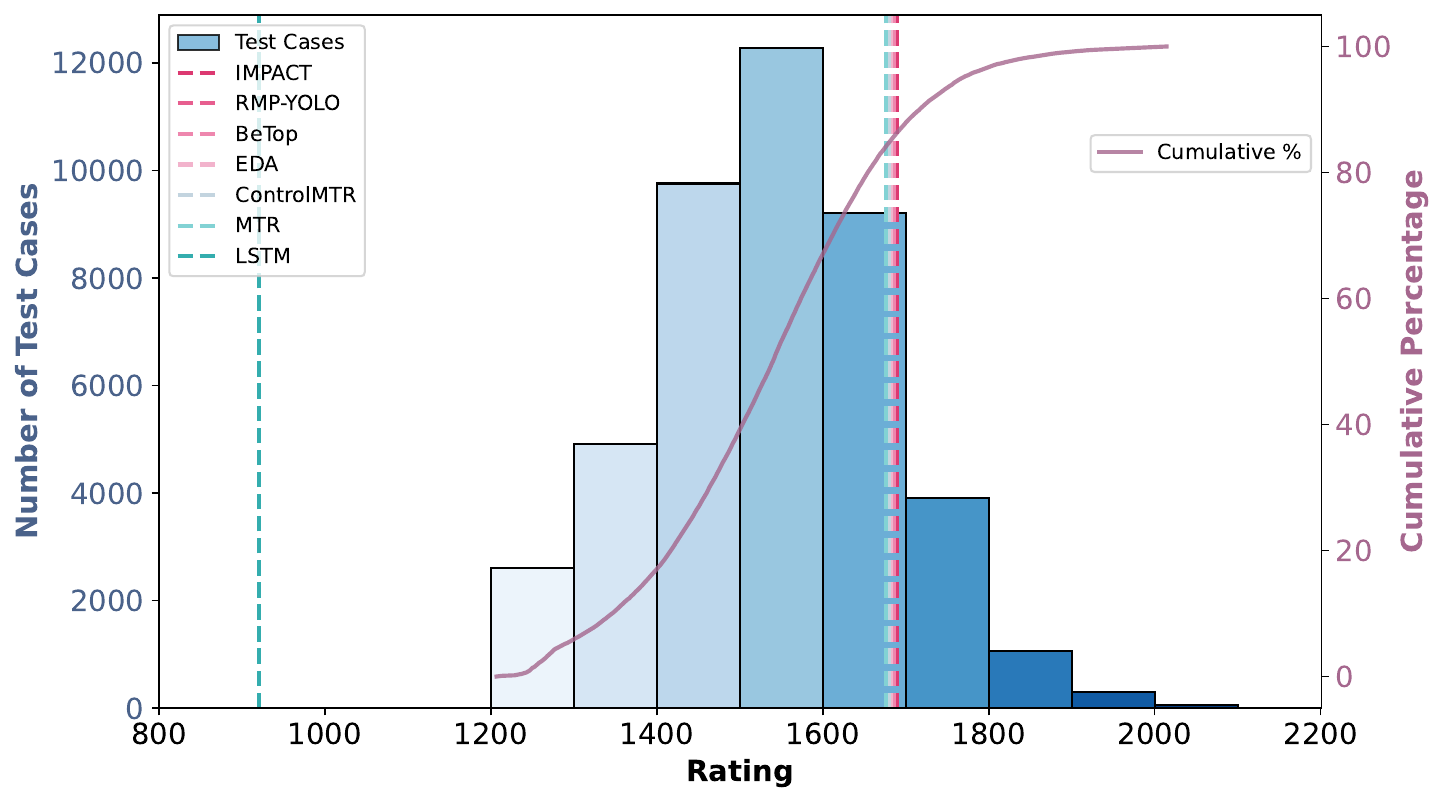}}
    \hfill
    \subfloat[Motion planning: NAVSIM~\cite{NAVSIM}]{\includegraphics[width=0.5\textwidth]{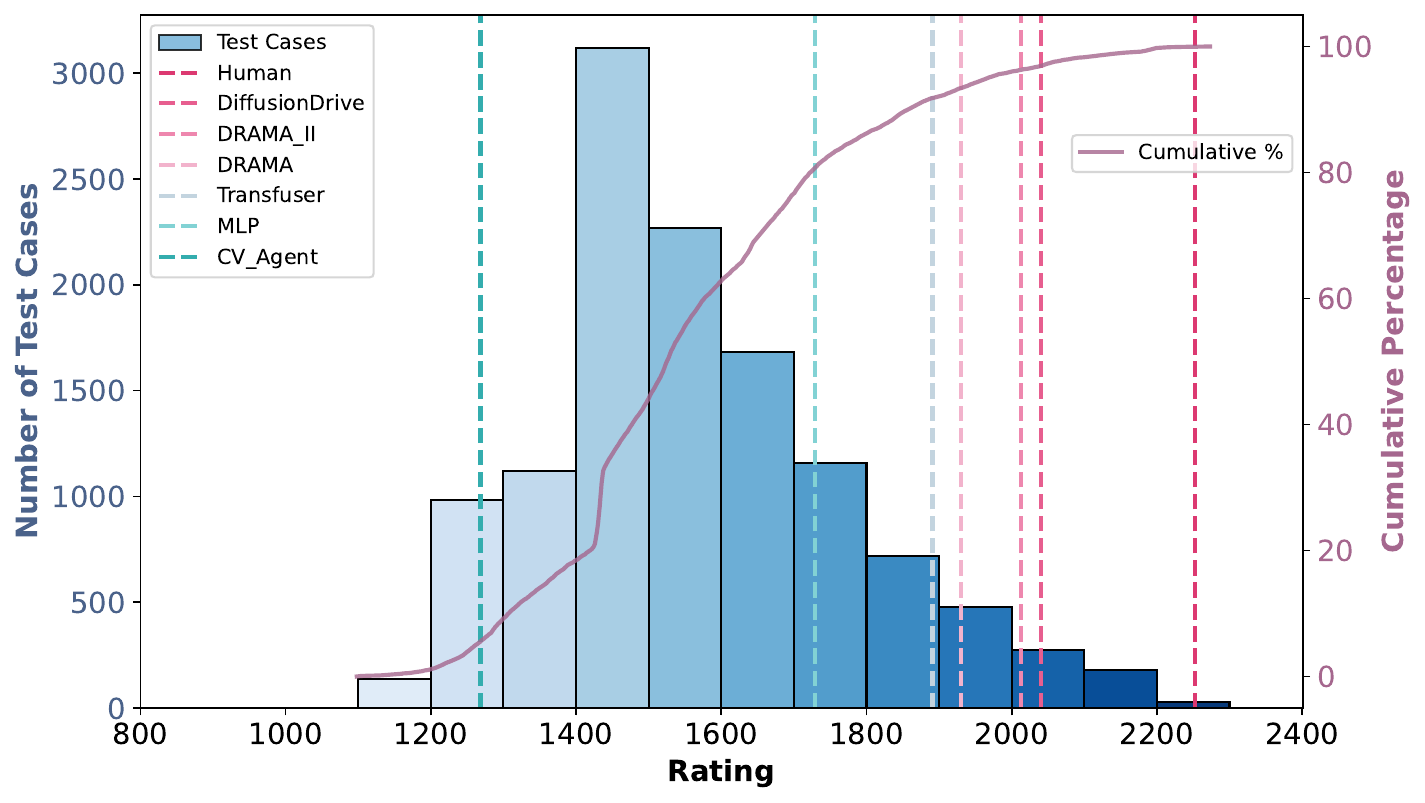}}
    
    \caption{
        Visualization of the estimated \textcolor{cyan}{test case} rating distribution and \textcolor{magenta}{agent} ratings on six distinct datasets. The \textcolor{purple}{percentile curve} represents the cumulative percentage of test cases up to each rating level. For each \textcolor{magenta}{agent}, the portion of the \textcolor{cyan}{test cases} and the \textcolor{purple}{percentile curve} that lies to the right represents the fraction of the dataset that remains difficult (below 50\% confidence).
    }
    \label{fig:rating_dist}
\end{figure}

In \autoref{fig:rating_dist}, the rating distributions of both test cases and agents are visualized across all six datasets. To provide a \textbf{qualitative evaluation} of test case difficulty, we randomly sample test cases from each rating level for every dataset/task and present them in \autoref{appendix:results} for visual comparison.


\begin{table}[htbp]
\centering
\caption{Competency gaps estimated on each dataset (excluding human)}
\label{tab:competency_gap}
\resizebox{0.99\textwidth}{!}{
    \begin{tabular}{cccccccccc} 
    \toprule
    \multirow{2}{*}{Domain} & \multirow{2}{*}{Task} & \multirow{2}{*}{Dataset} & \multirow{2}{*}{Metric} & \multirow{2}{*}{$R_{t, \text{max}}$} & \multirow{2}{*}{$R_{a, \text{max}}$} & \multirow{2}{*}{$\mathbb{E}[M_{a,t}] \uparrow$} & \multicolumn{3}{c}{Competency Gap to Oracles $\downarrow$} \\ 
    \cmidrule(lr){8-10}
     &  &  &  &  &  &  & @50\% & @90\% & @99\% \\ 
    \midrule
    \multirow{2}{*}{Vision} & Classification & ImageNet~\cite{deng2009imagenet} & Acc@1 & 2389.7 & 2035.0 & 0.115 & 354.7 & 736.4 & 1152.9 \\
     & Detection & COCO~\cite{mscoco} & AP@[.50:.90] & 2132.7 & 1745.5 & 0.097 & 387.2 & 768.9 & 1185.4 \\ 
    \cmidrule(lr){1-10}
    \multirow{2}{*}{Language} & QA & MMLU~\cite{MMLU} & Accuracy & 2446.1 & 2159.2 & 0.161 & 286.9 & 668.6 & 1085.1 \\
     & Coding & LiveCodeBench~\cite{jain2024livecodebench} & PassAll & 2263.3 & 1939.7 & 0.134 & 323.6 & 705.3 & 1121.8 \\ 
    \cmidrule(lr){1-10}
    \multirow{2}{*}{Action} & Prediction & Waymo~\cite{waymodataset} & mAP & 2014.3 & 1689.8 & 0.134 & 324.5 & 706.2 & 1122.8 \\
     & Planning & NAVSIM~\cite{NAVSIM}& PDM Score & 2273.0 & 2040.5 & \textbf{0.208} & \textbf{232.5} & \textbf{614.2} & \textbf{1030.8} \\
    \bottomrule
    \end{tabular}
}
\end{table}

From \autoref{fig:rating_dist}, we observe distinct test case difficulty distributions across different datasets by examining the histograms and the percentile curves over the rating spectrum. 
Datasets such as ImageNet~\cite{deng2009imagenet}, MMLU~\cite{MMLU}, and NAVSIM~\cite{NAVSIM} exhibit long-tail distributions, indicated by a small fraction of highly challenging test cases. In contrast,  LiveCodeBench~\cite{jain2024livecodebench} and Waymo~\cite{waymodataset} present more symmetrical distributions from the agents' perspectives, indicating a more balanced spread of difficulty levels. Meanwhile, COCO~\cite{mscoco} shows a short-tail distribution, suggesting that its most difficult test cases are relatively moderate in comparison.

By observing the improvements in model performance, we can trace the progress made on each task over the years. For example, on ImageNet~\cite{deng2009imagenet}, ConvNeXt-Large~\cite{liu2022convnet} (2022) obtained a rating of 2035, successfully surpassing approximately 85\% of test images (rated \(< 2035\)) with at least 50\% confidence, and about 67\% of images (rated \(< 1635 = 2035 - 400\)) with at least 91\% confidence. 
Compared to AlexNet~\cite{krizhevsky2012alexnet} (2012), which beats 64\% of the dataset with a rating of 1586, the progress over 10 years is about 449 rating points, and newly mastering 18\% of the dataset.

In \autoref{tab:competency_gap}, we report the highest-rated agents and test cases for each dataset, along with the predicted expected performance of each agent on the most difficult test case and the corresponding competency gaps to oracles at various confidence thresholds.

The results show that, excluding the human agent, the highest-rated AI models across the six datasets generally exhibit competency gaps of approximately 233–387 rating points from achieving mastery on the most difficult test cases at the 50\% confidence level, and approximately 1031–1185 rating points from oracles @ 99\% confidence level. In contrast, the human expert on the NAVSIM~\cite{NAVSIM} dataset achieves near-oracle-level competency under the PDM score metric, with a gap of only 20.7 rating points from the oracle @ 50\% confidence. This suggests that the human agent is approaching the performance of an ideal oracle on this task. 
These findings highlight that, in the presence of challenging test cases, current AI models remain significantly below oracle-level performance and face substantial competency gaps that must be bridged before achieving true task mastery.

\subsection{Reliability of the rating system}
\label{sec:reliability}

As the proposed method is uniquely designed for rating \textit{heterogeneous} players, it is essential to evaluate the reliability of the resultant ratings to ensure meaningful interpretations and to validate the assumptions underlying the design of the rating system. We assess rating reliability from two key perspectives: \textbf{consistency} with existing evaluation metrics and \textbf{predictive accuracy}.

\textbf{Consistency}: Spearman's rank correlation is used to measure the consistency between our estimated rating rankings and the original task-specific performance metrics. For each test case \(t\), we record the average agent performance \(\bar{M}_{t}\) on that test case, and for each agent \(a\), we compute the average agent performance \(\bar{M}_{a}\) across all test cases. The Spearman's rank correlation coefficient \(\rho_t\) between the rankings of \(\{R_t\}\) and \(\{\bar{M}_t\}\), and \(\rho_a\) between the rankings of \(\{R_a\}\) and \(\{\bar{M}_a\}\), are used as indicators.

\textbf{Predictive accuracy}: For each agent, its average performance \(\bar{M}_{a,B} = \frac{1}{|B|} \sum_{t \in B} M_{a,t}\) on all test cases within the same rating bin \( B \) is computed and compared against the theoretical expectations \( \mathbb{E}[M_{a,B}] \) derived from the rating system. 
The mean absolute error (MAE) and mean squared error (MSE) are used to quantify the deviation between the empirical performance \( \bar{M}_{a,B} \) and the theoretical expectation \( \mathbb{E}[M_{a,B}] \).

\begin{table}[htbp]
\centering
\caption{
    Consistency \& predictive accuracy across various datasets
}
\label{tab:reliability}
\resizebox{0.99\textwidth}{!}{
    \begin{tabular}{ccrrrcccc} 
    \toprule
    \multirow{2}{*}{Dataset} & \multirow{2}{*}{Split} & \multicolumn{1}{c}{\multirow{2}{*}{$N_{t}$}} & \multicolumn{1}{c}{\multirow{2}{*}{$N_{a}$}} & \multicolumn{1}{c}{\multirow{2}{*}{$N_{match}$}} & \multicolumn{2}{c}{Consistency} & \multicolumn{2}{c}{Predictive Accuracy} \\ 
    \cmidrule(r){6-9}
     &  & \multicolumn{1}{c}{} & \multicolumn{1}{c}{} & \multicolumn{1}{c}{} & $\rho_t \downarrow$ & $\rho_a \uparrow$ & MAE~$\downarrow$ & MSE$\downarrow$ \\ 
    \midrule
    ImageNet~\cite{deng2009imagenet} & val & 50,000 & 20 & 1,000,000 & -0.9685 & 0.9985 & 0.0476 & 0.0039 \\
    COCO~\cite{mscoco} & val & 4,952 & 20 & 99,040 & -0.9999 & 1.0000 & 0.0167 & 0.0005 \\
    MMLU~\cite{MMLU} & test & 13,957 & 20 & 279,140 & -0.9962 & 1.0000 & 0.0662 & 0.0076 \\
    LiveCodeBench~\cite{jain2024livecodebench} & test & 880 & 20 & 17,600 & -0.9968 & 0.9985 & 0.0446 & 0.0038 \\
    Waymo~\cite{waymodataset} & val & 44,097 & 7 & 308,679 & -0.9981 & 1.0000 & 0.0354 & 0.0023 \\
    NAVSIM~\cite{NAVSIM} & test & 12,147 & 7 & 85,050 & -0.9963 & 1.0000 & 0.0546 & 0.0088 \\
    \bottomrule
    \end{tabular}
}
\end{table}

\begin{figure}[htbp]
    \centering
    \includegraphics[width=0.99\textwidth]{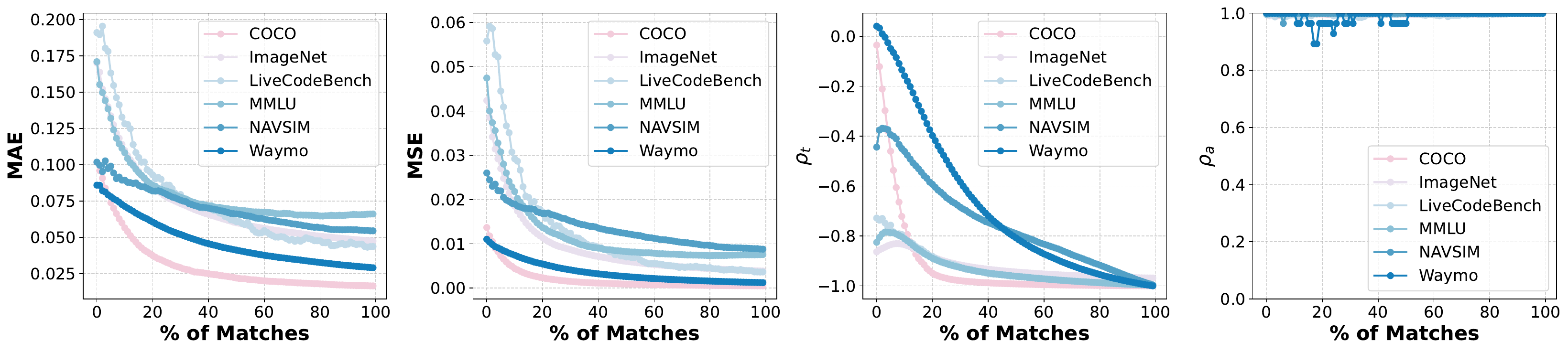}
    \caption{
        Evaluation of the reliability as a function of the percentage of completed matches.
    }
    \label{fig:eval_metrics}
\end{figure}

As shown in \autoref{tab:reliability}, the experimental results demonstrate that our method achieves consistently high correlation, indicating strong association between the derived ratings and the original evaluation metrics. Low MAE and MSE underscores the accuracy in predictive power. 

In \autoref{fig:eval_metrics}, we plot the evolution of all four evaluation metrics as a function of the percentage of matches used by the rating system. As more match data is incorporated, both MAE and MSE consistently decrease, indicating the convergence and stability of the system. Similarly, the correlation grows stronger with additional match information, demonstrating the effectiveness of our method in accurately rating both test cases and agents. These trends provide empirical support for the transitivity assumption introduced earlier.

\section{Related works}

\paragraph{Estimating per-instance difficulty}
Evaluating instance difficulties in datasets is an important yet understudied field~\cite{vodrahalli2018all, swayamdipta2020dataset}. 
Some methods rely on hand-crafted features like word overlap ~\cite{bengio2009curriculum}, input length ~\cite{spitkovsky2010baby, gururangan2018annotation}, or similarity scores~\cite{mishra2022hardness} as proxies for difficulty, which are oversimple. 
Many techniques adopt loss-based metrics~\cite{han2018co, arazo2019unsupervised, shen2019learning} or prediction confidence~\cite{hovy2013learning, chang2017active, varshney2022ildae}. 
Approaches like~\cite{toneva2018empirical, swayamdipta2020dataset, ethayarajh2022understanding} leverage model training dynamics, which can offer deeper insights, but are often influenced by the stochastic nature of training. 
However, these methods often yield model-specific difficulty estimates that are difficult to compare across models due to varying loss designs, and they are typically inapplicable to non-learning agents like classical algorithms or human agents. In contrast, our system directly utilizes performance metrics as difficulty indicators, making it broadly compatible and capable of capturing insights from a wide range of agents. This universality ensures that the estimated difficulties are meaningful and comparable across different agent types. 


\paragraph{Benchmarking AI capabilities}


Inspired by competitive games, several works have adopted rating systems to evaluate AI model performance across tasks or in head-to-head comparisons. For example, rating systems have been used to assess AlphaStar agents in StarCraft II competitions~\cite{vinyals2019grandmaster} and in reinforcement learning tournaments~\cite{herbrich2006trueskill}. The Chatbot Arena framework~\cite{chiang2024chatbot} applies a modified Elo system to conduct pairwise comparisons of large language models (LLMs), based on crowd-sourced human preference judgments.
However, these evaluation approaches typically focus solely on modeling agent capabilities, without accounting for the implicit difficulty of individual test cases. As a result, the estimated model ratings may fail to reflect true performance under varying levels of difficulty and can be unreliable~\cite{boubdir2024elo}. Furthermore, such model-vs-model competition setups are not easily generalizable to a wide range of AI tasks beyond dialogue or games.

Psychometric benchmarks~\cite{zhuang2023static, li2024quantifying} have also been applied to the AI domain to assess question difficulty and model ability. In particular, Item Response Theory (IRT) has been adapted to characterize the relative competency of models across tasks and datasets, enabling fine-grained performance profiling~\cite{martinez2019item}. However, prior work have primarily focused on basic machine learning tasks with simple classifiers, without extending to a broad range of complex tasks and state-of-the-art (SOTA) models.
By integrating rating systems with IRT-inspired evaluation, our framework offers a unified and interpretable approach to jointly estimate test case difficulty and model competency. This enables more reliable predictions for models on tasks, while preserving generalizability.

\section{Conclusion and limitations}

In this paper, we propose \textit{AGI-Elo}, a unified framework for jointly estimating task difficulties and agent competencies through a quantifiable, general-purpose rating system tailored for AGI tasks. Experimental results across six diverse tasks spanning vision, language, and action domains demonstrate the broad applicability and high predictive accuracy of our approach. The resulting rating distributions enable in-depth analysis of dataset difficulty characteristics, precise identification of long-tailed challenging test cases, and quantification of competency gaps between current AI agents and idealized oracles at various levels. To support further research, we release the computed test case and agent ratings, and we hope that our findings will stimulate broader interest in this important yet underexplored area.

While our results offer a novel perspective, they are not exhaustive. Due to limited computational resources, our current experimental scale is constrained, and the selected datasets and models may not fully represent state-of-the-art performance. Nevertheless, we believe the proposed methodology is sound, and we envision future studies expanding upon it with more comprehensive evaluations across the full spectrum of AGI capabilities.

\bibliographystyle{plain}
\bibliography{ref}

\begin{thebibliography}{10}

\bibitem{achiam2023gpt}
Josh Achiam, Steven Adler, Sandhini Agarwal, Lama Ahmad, Ilge Akkaya, Florencia~Leoni Aleman, Diogo Almeida, Janko Altenschmidt, Sam Altman, Shyamal Anadkat, et~al.
\newblock Gpt-4 technical report.
\newblock {\em arXiv preprint arXiv:2303.08774}, 2023.

\bibitem{claudesonnet}
{Anthropic}.
\newblock Claude 3.7 sonnet and claude code.
\newblock \url{https://www.anthropic.com/news/claude-3-7-sonnet}, February 2025.
\newblock Accessed: 2025-05-16.

\bibitem{arazo2019unsupervised}
Eric Arazo, Daniel Ortego, Paul Albert, Noel~E O'Connor, and Kevin McGuinness.
\newblock Unsupervised label noise modeling and loss correction.
\newblock In {\em International Conference on Machine Learning}, pages 312--321, 2019.

\bibitem{bengio2009curriculum}
Yoshua Bengio, Jerome Louradour, Ronan Collobert, and Jason Weston.
\newblock Curriculum learning.
\newblock In {\em Proceedings of the 26th annual international conference on machine learning}, pages 41--48, 2009.

\bibitem{boubdir2024elo}
Meriem Boubdir, Edward Kim, Beyza Ermis, Sara Hooker, and Marzieh Fadaee.
\newblock Elo uncovered: Robustness and best practices in language model evaluation.
\newblock In {\em The Thirty-eighth Annual Conference on Neural Information Processing Systems}, 2024.

\bibitem{nuscenes}
Holger Caesar, Varun Bankiti, Alex~H. Lang, Sourabh Vora, Venice~Erin Liong, Qiang Xu, Anush Krishnan, Yu~Pan, Giancarlo Baldan, and Oscar Beijbom.
\newblock nuscenes: A multimodal dataset for autonomous driving.
\newblock In {\em CVPR}, 2020.

\bibitem{chang2017active}
Hwanjun~Songkuk Chang and et~al.
\newblock Active bias: Training more accurate neural networks by emphasizing high variance samples.
\newblock In {\em Advances in Neural Information Processing Systems}, pages 1002--1012, 2017.

\bibitem{chen2019mmdetection}
Kai Chen, Jiaqi Wang, Jiangmiao Pang, Yuhang Cao, Yu~Xiong, Xiaoxiao Li, Shuyang Sun, Wansen Feng, Ziwei Liu, Jiarui Xu, et~al.
\newblock Mmdetection: Open mmlab detection toolbox and benchmark.
\newblock {\em arXiv preprint arXiv:1906.07155}, 2019.

\bibitem{chiang2024chatbot}
Wei-Lin Chiang, Lianmin Zheng, Ying Sheng, Anastasios~Nikolas Angelopoulos, Tianle Li, Dacheng Li, Banghua Zhu, Hao Zhang, Michael Jordan, Joseph~E Gonzalez, et~al.
\newblock Chatbot arena: An open platform for evaluating llms by human preference.
\newblock In {\em Forty-first International Conference on Machine Learning}, 2024.

\bibitem{Chitta2023PAMI}
Kashyap Chitta, Aditya Prakash, Bernhard Jaeger, Zehao Yu, Katrin Renz, and Andreas Geiger.
\newblock Transfuser: Imitation with transformer-based sensor fusion for autonomous driving.
\newblock {\em Pattern Analysis and Machine Intelligence (PAMI)}, 2023.

\bibitem{contributors2023openscene}
OpenScene Contributors.
\newblock Openscene: The largest up-to-date 3d occupancy prediction benchmark in autonomous driving, 2023.

\bibitem{NAVSIM}
Daniel Dauner, Marcel Hallgarten, Tianyu Li, Xinshuo Weng, Zhiyu Huang, Zetong Yang, Hongyang Li, Igor Gilitschenski, Boris Ivanovic, Marco Pavone, Andreas Geiger, and Kashyap Chitta.
\newblock Navsim: Data-driven non-reactive autonomous vehicle simulation and benchmarking.
\newblock In {\em Advances in Neural Information Processing Systems (NeurIPS)}, 2024.

\bibitem{deng2009imagenet}
Jia Deng, Wei Dong, Richard Socher, Li-Jia Li, Kai Li, and Li~Fei-Fei.
\newblock Imagenet: A large-scale hierarchical image database.
\newblock In {\em Proceedings of the IEEE Conference on Computer Vision and Pattern Recognition (CVPR)}, pages 248--255, 2009.

\bibitem{dosovitskiy2020image}
Alexey Dosovitskiy, Lucas Beyer, Alexander Kolesnikov, Dirk Weissenborn, Xiaohua Zhai, Thomas Unterthiner, Mostafa Dehghani, Matthias Minderer, Georg Heigold, Sylvain Gelly, et~al.
\newblock An image is worth 16x16 words: Transformers for image recognition at scale.
\newblock {\em arXiv preprint arXiv:2010.11929}, 2020.

\bibitem{elo1967proposed}
Arpad~E Elo.
\newblock The proposed uscf rating system, its development, theory, and applications.
\newblock {\em Chess life}, 22(8):242--247, 1967.

\bibitem{ethayarajh2022understanding}
Kawin Ethayarajh, Yejin Choi, and Swabha Swayamdipta.
\newblock Understanding dataset difficulty with v-usable information.
\newblock In {\em International Conference on Machine Learning}, pages 5988--6008. PMLR, 2022.

\bibitem{waymodataset}
Scott Ettinger, Shuyang Cheng, Benjamin Caine, Chenxi Liu, Hang Zhao, Sabeek Pradhan, Yuning Chai, Ben Sapp, Charles~R Qi, Yin Zhou, et~al.
\newblock Large scale interactive motion forecasting for autonomous driving: The waymo open motion dataset.
\newblock In {\em Proceedings of the IEEE/CVF International Conference on Computer Vision}, pages 9710--9719, 2021.

\bibitem{Geiger2012CVPR-kitti}
Andreas Geiger, Philip Lenz, and Raquel Urtasun.
\newblock Are we ready for autonomous driving? the kitti vision benchmark suite.
\newblock In {\em Conference on Computer Vision and Pattern Recognition (CVPR)}, 2012.

\bibitem{glickman1999glicko}
Mark~E. Glickman.
\newblock Parameter estimation in large dynamic paired comparison experiments.
\newblock {\em Journal of the Royal Statistical Society: Series C (Applied Statistics)}, 48(3):377--394, 1999.

\bibitem{google2025gemini}
{Google DeepMind}.
\newblock Gemini 2.5: Our most intelligent ai model.
\newblock \url{https://blog.google/technology/google-deepmind/gemini-model-thinking-updates-march-2025/}, March 2025.
\newblock Accessed: 2025-05-16.

\bibitem{grattafiori2024llama}
Aaron Grattafiori, Abhimanyu Dubey, Abhinav Jauhri, Abhinav Pandey, Abhishek Kadian, Ahmad Al-Dahle, Aiesha Letman, Akhil Mathur, Alan Schelten, Alex Vaughan, et~al.
\newblock The llama 3 herd of models.
\newblock {\em arXiv preprint arXiv:2407.21783}, 2024.

\bibitem{guo2025deepseek}
Daya Guo, Dejian Yang, Haowei Zhang, Junxiao Song, Ruoyu Zhang, Runxin Xu, Qihao Zhu, Shirong Ma, Peiyi Wang, Xiao Bi, et~al.
\newblock Deepseek-r1: Incentivizing reasoning capability in llms via reinforcement learning.
\newblock {\em arXiv preprint arXiv:2501.12948}, 2025.

\bibitem{guo2024deepseek}
Daya Guo, Qihao Zhu, Dejian Yang, Zhenda Xie, Kai Dong, Wentao Zhang, Guanting Chen, Xiao Bi, Yu~Wu, YK~Li, et~al.
\newblock Deepseek-coder: When the large language model meets programming--the rise of code intelligence.
\newblock {\em arXiv preprint arXiv:2401.14196}, 2024.

\bibitem{gururangan2018annotation}
Suchin Gururangan, Swabha Swayamdipta, Omer Levy, Roy Schwartz, Samuel Bowman, and Noah~A. Smith.
\newblock Annotation artifacts in natural language inference data.
\newblock In {\em Proceedings of the 2018 Conference of the North American Chapter of the Association for Computational Linguistics: Human Language Technologies}, volume~2, pages 107--112, 2018.

\bibitem{nuplan}
K.~Tan et~al. H.~Caesar, J.~Kabzan.
\newblock Nuplan: A closed-loop ml-based planning benchmark for autonomous vehicles.
\newblock In {\em CVPR ADP3 workshop}, 2021.

\bibitem{han2018co}
Bo~Han, Quanming Yao, Xingrui Yu, Gang Niu, Miao Xu, Weihua Hu, Ivor~W Tsang, and Masashi Sugiyama.
\newblock Co-teaching: Robust training of deep neural networks with extremely noisy labels.
\newblock In {\em Advances in Neural Information Processing Systems}, pages 8527--8537, 2018.

\bibitem{OD-MaskRCNN}
Kaiming He, Georgia Gkioxari, Piotr Doll{\'a}r, and Ross Girshick.
\newblock Mask r-cnn.
\newblock {\em arXiv preprint arXiv:1703.06870}, 2017.

\bibitem{he2016deep}
Kaiming He, Xiangyu Zhang, Shaoqing Ren, and Jian Sun.
\newblock Deep residual learning for image recognition.
\newblock In {\em Proceedings of the IEEE/CVF Conference on Computer Vision and Pattern Recognition (CVPR)}, pages 770--778, 2016.

\bibitem{MMLU}
Dan Hendrycks, Collin Burns, Steven Basart, Andy Zou, Mantas Mazeika, Dawn Song, and Jacob Steinhardt.
\newblock Measuring massive multitask language understanding.
\newblock {\em arXiv preprint arXiv:2009.03300}, 2020.

\bibitem{herbrich2006trueskill}
Ralf Herbrich, Tom Minka, and Thore Graepel.
\newblock Trueskill™: a bayesian skill rating system.
\newblock {\em Advances in neural information processing systems}, 19, 2006.

\bibitem{hovy2013learning}
Dirk Hovy, Barbara Plank, and Anders Sogaard.
\newblock Learning whodunnit: Classification of event participants in news articles.
\newblock In {\em Proceedings of the 2013 Conference on Empirical Methods in Natural Language Processing}, pages 540--545, 2013.

\bibitem{huang2017densely}
Gao Huang, Zhuang Liu, Laurens Van Der~Maaten, and Kilian~Q Weinberger.
\newblock Densely connected convolutional networks.
\newblock In {\em Proceedings of the IEEE conference on computer vision and pattern recognition}, pages 4700--4708, 2017.

\bibitem{hui2024qwen2}
Binyuan Hui, Jian Yang, Zeyu Cui, Jiaxi Yang, Dayiheng Liu, Lei Zhang, Tianyu Liu, Jiajun Zhang, Bowen Yu, Keming Lu, et~al.
\newblock Qwen2. 5-coder technical report.
\newblock {\em arXiv preprint arXiv:2409.12186}, 2024.

\bibitem{hurst2024gpt}
Aaron Hurst, Adam Lerer, Adam~P Goucher, Adam Perelman, Aditya Ramesh, Aidan Clark, AJ~Ostrow, Akila Welihinda, Alan Hayes, Alec Radford, et~al.
\newblock Gpt-4o system card.
\newblock {\em arXiv preprint arXiv:2410.21276}, 2024.

\bibitem{iandola2016squeezenet}
Forrest~N Iandola, Song Han, Matthew~W Moskewicz, Khalid Ashraf, William~J Dally, and Kurt Keutzer.
\newblock Squeezenet: Alexnet-level accuracy with 50x fewer parameters and< 0.5 mb model size.
\newblock {\em arXiv preprint arXiv:1602.07360}, 2016.

\bibitem{jaech2024openai}
Aaron Jaech, Adam Kalai, Adam Lerer, Adam Richardson, Ahmed El-Kishky, Aiden Low, Alec Helyar, Aleksander Madry, Alex Beutel, Alex Carney, et~al.
\newblock Openai o1 system card.
\newblock {\em arXiv preprint arXiv:2412.16720}, 2024.

\bibitem{jain2024livecodebench}
Naman Jain, King Han, Alex Gu, Wen-Ding Li, Fanjia Yan, Tianjun Zhang, Sida Wang, Armando Solar-Lezama, Koushik Sen, and Ion Stoica.
\newblock Livecodebench: Holistic and contamination free evaluation of large language models for code.
\newblock {\em arXiv preprint arXiv:2403.07974}, 2024.

\bibitem{yolov8}
Glenn Jocher, Ayush Chaurasia, and Jing Qiu.
\newblock Ultralytics yolov8, 2023.

\bibitem{yolov11}
Glenn Jocher and Jing Qiu.
\newblock Ultralytics yolo11, 2024.

\bibitem{jocher2020ultralytics}
Glenn Jocher, Alex Stoken, Jirka Borovec, Liu Changyu, Adam Hogan, Laurentiu Diaconu, Jake Poznanski, Lijun Yu, Prashant Rai, Russ Ferriday, et~al.
\newblock ultralytics/yolov5: v3. 0.
\newblock {\em Zenodo}, 2020.

\bibitem{Krizhevsky09-cifar}
Alex Krizhevsky.
\newblock Learning multiple layers of features from tiny images.
\newblock Technical report, 2009.

\bibitem{krizhevsky2012alexnet}
Alex Krizhevsky, Ilya Sutskever, and Geoffrey~E. Hinton.
\newblock Imagenet classification with deep convolutional neural networks.
\newblock In {\em Advances in Neural Information Processing Systems (NeurIPS)}, volume~25, pages 1097--1105, 2012.

\bibitem{lhoest2021datasets}
Quentin Lhoest, Albert~Villanova Del~Moral, Yacine Jernite, Abhishek Thakur, Patrick Von~Platen, Suraj Patil, Julien Chaumond, Mariama Drame, Julien Plu, Lewis Tunstall, et~al.
\newblock Datasets: A community library for natural language processing.
\newblock {\em arXiv preprint arXiv:2109.02846}, 2021.

\bibitem{li2024quantifying}
Yuan Li, Yue Huang, Hongyi Wang, Xiangliang Zhang, James Zou, and Lichao Sun.
\newblock Quantifying ai psychology: A psychometrics benchmark for large language models.
\newblock {\em arXiv preprint arXiv:2406.17675}, 2024.

\bibitem{diffusiondrive}
Bencheng Liao, Shaoyu Chen, Haoran Yin, Bo~Jiang, Cheng Wang, Sixu Yan, Xinbang Zhang, Xiangyu Li, Ying Zhang, Qian Zhang, and Xinggang Wang.
\newblock Diffusiondrive: Truncated diffusion model for end-to-end autonomous driving.
\newblock {\em arXiv preprint arXiv:2411.15139}, 2024.

\bibitem{lin2024eda}
Longzhong Lin, Xuewu Lin, Tianwei Lin, Lichao Huang, Rong Xiong, and Yue Wang.
\newblock Eda: Evolving and distinct anchors for multimodal motion prediction.
\newblock In {\em Proceedings of the AAAI Conference on Artificial Intelligence}, volume~38, pages 3432--3440, 2024.

\bibitem{OD-RetinaNet}
Tsung-Yi Lin, Priya Goyal, Ross Girshick, Kaiming He, and Piotr Doll{\'a}r.
\newblock Focal loss for dense object detection.
\newblock {\em arXiv preprint arXiv:1708.02002}, 2017.

\bibitem{mscoco}
Tsung-Yi Lin, Michael Maire, Serge Belongie, James Hays, Pietro Perona, Deva Ramanan, Piotr Doll{\'a}r, and C~Lawrence Zitnick.
\newblock Microsoft coco: Common objects in context.
\newblock In {\em Computer vision--ECCV 2014: 13th European conference, zurich, Switzerland, September 6-12, 2014, proceedings, part v 13}, pages 740--755. Springer, 2014.

\bibitem{liu2024deepseek}
Aixin Liu, Bei Feng, Bing Xue, Bingxuan Wang, Bochao Wu, Chengda Lu, Chenggang Zhao, Chengqi Deng, Chenyu Zhang, Chong Ruan, et~al.
\newblock Deepseek-v3 technical report.
\newblock {\em arXiv preprint arXiv:2412.19437}, 2024.

\bibitem{liu2024betop}
Haochen Liu, Li~Chen, Yu~Qiao, Chen Lv, and Hongyang Li.
\newblock Reasoning multi-agent behavioral topology for interactive autonomous driving.
\newblock In {\em NeurIPS}, 2024.

\bibitem{OD-SSD-VGG16}
Wei Liu, Dragomir Anguelov, Dumitru Erhan, Christian Szegedy, Scott Reed, Cheng-Yang Fu, and Alexander~C Berg.
\newblock Ssd: Single shot multibox detector.
\newblock In {\em Computer Vision--ECCV 2016: 14th European Conference, Amsterdam, The Netherlands, October 11--14, 2016, Proceedings, Part I 14}, pages 21--37. Springer, 2016.

\bibitem{liu2021swin}
Ze~Liu, Yutong Lin, Yue Cao, Han Hu, Yixuan Wei, Zheng Zhang, Stephen Lin, and Baining Guo.
\newblock Swin transformer: Hierarchical vision transformer using shifted windows.
\newblock In {\em Proceedings of the IEEE/CVF International Conference on Computer Vision (ICCV)}, pages 10012--10022, 2021.

\bibitem{liu2022convnet}
Zhuang Liu, Hanzi Mao, Chao-Yuan Wu, Christoph Feichtenhofer, Trevor Darrell, and Saining Xie.
\newblock A convnet for the 2020s.
\newblock In {\em Proceedings of the IEEE/CVF conference on computer vision and pattern recognition}, pages 11976--11986, 2022.

\bibitem{lord2008statistical}
Frederic~M Lord and Melvin~R Novick.
\newblock {\em Statistical theories of mental test scores}.
\newblock IAP, 2008.

\bibitem{lozhkov2024starcoder}
Anton Lozhkov, Raymond Li, Loubna~Ben Allal, Federico Cassano, Joel Lamy-Poirier, Nouamane Tazi, Ao~Tang, Dmytro Pykhtar, Jiawei Liu, Yuxiang Wei, et~al.
\newblock Starcoder 2 and the stack v2: The next generation.
\newblock {\em arXiv preprint arXiv:2402.19173}, 2024.

\bibitem{baidu-RT-DETR}
Wenyu Lv, Shangliang Xu, Yian Zhao, Guanzhong Wang, Jinman Wei, Cheng Cui, Yuning Du, Qingqing Dang, and Yi~Liu.
\newblock Detrs beat yolos on real-time object detection (2023).
\newblock {\em arXiv preprint arXiv:2304.08069}, 2023.

\bibitem{ma2018shufflenet}
Ningning Ma, Xiangyu Zhang, Hai-Tao Zheng, and Jian Sun.
\newblock Shufflenet v2: Practical guidelines for efficient cnn architecture design.
\newblock In {\em Proceedings of the European conference on computer vision (ECCV)}, pages 116--131, 2018.

\bibitem{martinez2019item}
Fernando Mart{\'\i}nez-Plumed, Ricardo~BC Prud{\^e}ncio, Adolfo Mart{\'\i}nez-Us{\'o}, and Jos{\'e} Hern{\'a}ndez-Orallo.
\newblock Item response theory in ai: Analysing machine learning classifiers at the instance level.
\newblock {\em Artificial intelligence}, 271:18--42, 2019.

\bibitem{meta2025llama4}
{Meta AI}.
\newblock The llama 4 herd: The beginning of a new era of natively multimodal ai innovation.
\newblock \url{https://ai.meta.com/blog/llama-4-multimodal-intelligence/}, April 2025.
\newblock Accessed: 2025-05-16.

\bibitem{mishra2022hardness}
Swaroop Mishra, Anjana Arunkumar, Chris Bryan, and Chitta Baral.
\newblock Hardness of samples need to be quantified for a reliable evaluation system: Exploring potential opportunities with a new task.
\newblock {\em arXiv preprint arXiv:2210.07631}, 2022.

\bibitem{muennighoff2022crosslingual}
Niklas Muennighoff, Thomas Wang, Lintang Sutawika, Adam Roberts, Stella Biderman, Teven~Le Scao, M~Saiful Bari, Sheng Shen, Zheng-Xin Yong, Hailey Schoelkopf, et~al.
\newblock Crosslingual generalization through multitask finetuning.
\newblock {\em arXiv preprint arXiv:2211.01786}, 2022.

\bibitem{nanogpt}
NanoGPT.
\newblock Nanogpt api.
\newblock \url{https://nano-gpt.com/api}, 2025.
\newblock Accessed: 2025-05-15.

\bibitem{openaiAPI}
{OpenAI}.
\newblock Openai api.
\newblock \url{https://platform.openai.com}, 2025.
\newblock Accessed: 2025-05-15.

\bibitem{paszke2019pytorch}
A~Paszke.
\newblock Pytorch: An imperative style, high-performance deep learning library.
\newblock {\em arXiv preprint arXiv:1912.01703}, 2019.

\bibitem{qwen2024qwq32b}
{Qwen Team}.
\newblock Qwq: Reflect deeply on the boundaries of the unknown.
\newblock \url{https://qwenlm.github.io/blog/qwq-32b-preview/}, November 2024.
\newblock Accessed: 2025-05-16.

\bibitem{radosavovic2020designing}
Ilija Radosavovic, Raj~Prateek Kosaraju, Ross Girshick, Kaiming He, and Piotr Doll{\'a}r.
\newblock Designing network design spaces.
\newblock In {\em Proceedings of the IEEE/CVF conference on computer vision and pattern recognition}, pages 10428--10436, 2020.

\bibitem{yolov3}
Joseph Redmon and Ali Farhadi.
\newblock Yolov3: An incremental improvement.
\newblock {\em arXiv preprint arXiv:1804.02767}, 2018.

\bibitem{OD-FasterRCNN}
Shaoqing Ren, Kaiming He, Ross Girshick, and Jian Sun.
\newblock Faster r-cnn: Towards real-time object detection with region proposal networks.
\newblock {\em arXiv preprint arXiv:1506.01497}, 2015.

\bibitem{sandler2018mobilenetv2}
Mark Sandler, Andrew Howard, Menglong Zhu, Andrey Zhmoginov, and Liang-Chieh Chen.
\newblock Mobilenetv2: Inverted residuals and linear bottlenecks.
\newblock In {\em Proceedings of the IEEE/CVF Conference on Computer Vision and Pattern Recognition (CVPR)}, pages 4510--4520, 2018.

\bibitem{shen2019learning}
Yao Shen and Sujay Sanghavi.
\newblock Learning with bad training data via iterative trimmed loss minimization.
\newblock In {\em International Conference on Machine Learning}, pages 5739--5748, 2019.

\bibitem{mtr}
Shaoshuai Shi, Li~Jiang, Dengxin Dai, and Bernt Schiele.
\newblock Motion transformer with global intention localization and local movement refinement.
\newblock {\em Advances in Neural Information Processing Systems}, 2022.

\bibitem{simonyan2014very}
Karen Simonyan and Andrew Zisserman.
\newblock Very deep convolutional networks for large-scale image recognition.
\newblock {\em arXiv preprint arXiv:1409.1556}, 2014.

\bibitem{spitkovsky2010baby}
Valentin~I Spitkovsky, Hiyan Alshawi, and Dan Jurafsky.
\newblock Baby steps: How "less is more" in unsupervised dependency parsing.
\newblock In {\em Human Language Technologies: The 2010 Annual Conference of the North American Chapter of the Association for Computational Linguistics}, pages 751--759, 2010.

\bibitem{sun2024rmp}
Jiawei Sun, Jiahui Li, Tingchen Liu, Chengran Yuan, Shuo Sun, Zefan Huang, Anthony Wong, Keng~Peng Tee, and Marcelo~H Ang~Jr.
\newblock Rmp-yolo: A robust motion predictor for partially observable scenarios even if you only look once.
\newblock {\em arXiv preprint arXiv:2409.11696}, 2024.

\bibitem{controlmtr}
Jiawei Sun, Chengran Yuan, Shuo Sun, Shanze Wang, Yuhang Han, Shuailei Ma, Zefan Huang, Anthony Wong, Keng~Peng Tee, and Marcelo~H. Ang.
\newblock Controlmtr: Control-guided motion transformer with scene-compliant intention points for feasible motion prediction.
\newblock In {\em 2024 IEEE 27th International Conference on Intelligent Transportation Systems (ITSC)}, pages 1507--1514, 2024.

\bibitem{sun2025impact}
Jiawei Sun, Xibin Yue, Jiahui Li, Tianle Shen, Chengran Yuan, Shuo Sun, Sheng Guo, Quanyun Zhou, and Marcelo~H Ang~Jr.
\newblock Impact: Behavioral intention-aware multimodal trajectory prediction with adaptive context trimming.
\newblock {\em arXiv preprint arXiv:2504.09103}, 2025.

\bibitem{swayamdipta2020dataset}
Swabha Swayamdipta, Roy Schwartz, Nicholas Lourie, Yizhong Wang, Hannaneh Hajishirzi, Noah~A Smith, and Yejin Choi.
\newblock Dataset cartography: Mapping and diagnosing datasets with training dynamics.
\newblock {\em arXiv preprint arXiv:2009.10795}, 2020.

\bibitem{szegedy2016rethinking}
Christian Szegedy, Vincent Vanhoucke, Sergey Ioffe, Jon Shlens, and Zbigniew Wojna.
\newblock Rethinking the inception architecture for computer vision.
\newblock In {\em Proceedings of the IEEE conference on computer vision and pattern recognition}, pages 2818--2826, 2016.

\bibitem{tan2019efficientnet}
Mingxing Tan and Quoc~V. Le.
\newblock Efficientnet: Rethinking model scaling for convolutional neural networks.
\newblock In {\em Proceedings of the IEEE/CVF Conference on Computer Vision and Pattern Recognition (CVPR)}, pages 6105--6114, 2019.

\bibitem{team2025gemma}
Gemma Team, Aishwarya Kamath, Johan Ferret, Shreya Pathak, Nino Vieillard, Ramona Merhej, Sarah Perrin, Tatiana Matejovicova, Alexandre Ram{\'e}, Morgane Rivi{\`e}re, et~al.
\newblock Gemma 3 technical report.
\newblock {\em arXiv preprint arXiv:2503.19786}, 2025.

\bibitem{OD-FCOS}
Zhi Tian, Chunhua Shen, Hao Chen, and Tong He.
\newblock Fcos: Fully convolutional one-stage object detection.
\newblock In {\em Proceedings of the IEEE/CVF International Conference on Computer Vision (ICCV)}, pages 9627--9636, 2019.

\bibitem{toneva2018empirical}
Mariya Toneva, Alessandro Sordoni, Yulia Tsvetkov, Tommi Jaakkola, and Ellie Pavlick.
\newblock An empirical study of example forgetting during deep neural network learning.
\newblock In {\em International Conference on Learning Representations}, 2019.

\bibitem{varshney2022ildae}
Neeraj Varshney, Swaroop Mishra, and Chitta Baral.
\newblock Ildae: Instance-level difficulty analysis of evaluation data.
\newblock {\em arXiv preprint arXiv:2203.03073}, 2022.

\bibitem{vinyals2019grandmaster}
Oriol Vinyals, Igor Babuschkin, Wojciech~M Czarnecki, et~al.
\newblock Grandmaster level in starcraft ii using multi-agent reinforcement learning.
\newblock {\em Nature}, 575(7782):350--354, 2019.

\bibitem{vodrahalli2018all}
Kumar Vodrahalli, Ganesh Ramakrishnan, and Balaraman Ravindran.
\newblock Are all training examples created equal? an empirical study.
\newblock In {\em arXiv preprint arXiv:1803.07156}, 2018.

\bibitem{longtaillearning2024wang}
Haohui Wang, Weijie Guan, Jianpeng Chen, Zi~Wang, and Dawei Zhou.
\newblock Towards heterogeneous long-tailed learning: Benchmarking, metrics, and toolbox.
\newblock In {\em Advances in Neural Information Processing Systems 37 (NeurIPS 2024), Datasets and Benchmarks Track}, 2024.

\bibitem{xie2017aggregated}
Saining Xie, Ross Girshick, Piotr Doll{\'a}r, Zhuowen Tu, and Kaiming He.
\newblock Aggregated residual transformations for deep neural networks.
\newblock In {\em Proceedings of the IEEE conference on computer vision and pattern recognition}, pages 1492--1500, 2017.

\bibitem{yang2024qwen2}
An~Yang, Baosong Yang, Beichen Zhang, Binyuan Hui, Bo~Zheng, Bowen Yu, Chengyuan Li, Dayiheng Liu, Fei Huang, Haoran Wei, et~al.
\newblock Qwen2. 5 technical report.
\newblock {\em arXiv preprint arXiv:2412.15115}, 2024.

\bibitem{yuan2024drama}
Chengran Yuan, Zhanqi Zhang, Jiawei Sun, Shuo Sun, Zefan Huang, Christina Dao~Wen Lee, Dongen Li, Yuhang Han, Anthony Wong, Keng~Peng Tee, et~al.
\newblock Drama: An efficient end-to-end motion planner for autonomous driving with mamba.
\newblock {\em arXiv preprint arXiv:2408.03601}, 2024.

\bibitem{OD-ResNeSt}
Hang Zhang, Chongruo Wu, Zhongyue Zhang, Yi~Zhu, Zhi Zhang, Haibin Lin, Yue Sun, Tong He, Jonas Muller, R.~Manmatha, Mu~Li, and Alexander Smola.
\newblock Resnest: Split-attention networks.
\newblock {\em arXiv preprint arXiv:2004.08955}, 2020.

\bibitem{OD-DINO}
Hao Zhang, Feng Li, Shilong Liu, Lei Zhang, Hang Su, Jun Zhu, Lionel~M Ni, and Heung-Yeung Shum.
\newblock Dino: Detr with improved denoising anchor boxes for end-to-end object detection.
\newblock {\em arXiv preprint arXiv:2203.03605}, 2022.

\bibitem{zhuang2023static}
Yan Zhuang, Qi~Liu, Yuting Ning, Weizhe Huang, Zachary~A Pardos, Patrick~C Kyllonen, Jiyun Zu, Qingyang Mao, Rui Lv, Zhenya Huang, et~al.
\newblock From static benchmarks to adaptive testing: Psychometrics in ai evaluation.
\newblock {\em arXiv preprint arXiv:2306.10512}, 2023.

\end{thebibliography}


\appendix

\newpage

\section{Supplementary results for experiments}
\label{appendix:results}

\subsection{Qualitative evaluation} 
\label{appendix:results:qualitative}


We have made our qualitative examples available on:

\textbf{HuggingFace}:

\url{https://huggingface.co/collections/ztony0712/agi-elo-6825d88e9587700e9dd41b12}

\textbf{Project page}: 

\url{https://ss47816.github.io/AGI-Elo/}







\subsection{Performance prediction vs. reality: predictive accuracy on various datasets}
\label{appendix:results:reliability}

\begin{figure}[h]
    \centering
    \subfloat[Image classification: ImageNet~\cite{deng2009imagenet}]{\includegraphics[width=0.5\textwidth]{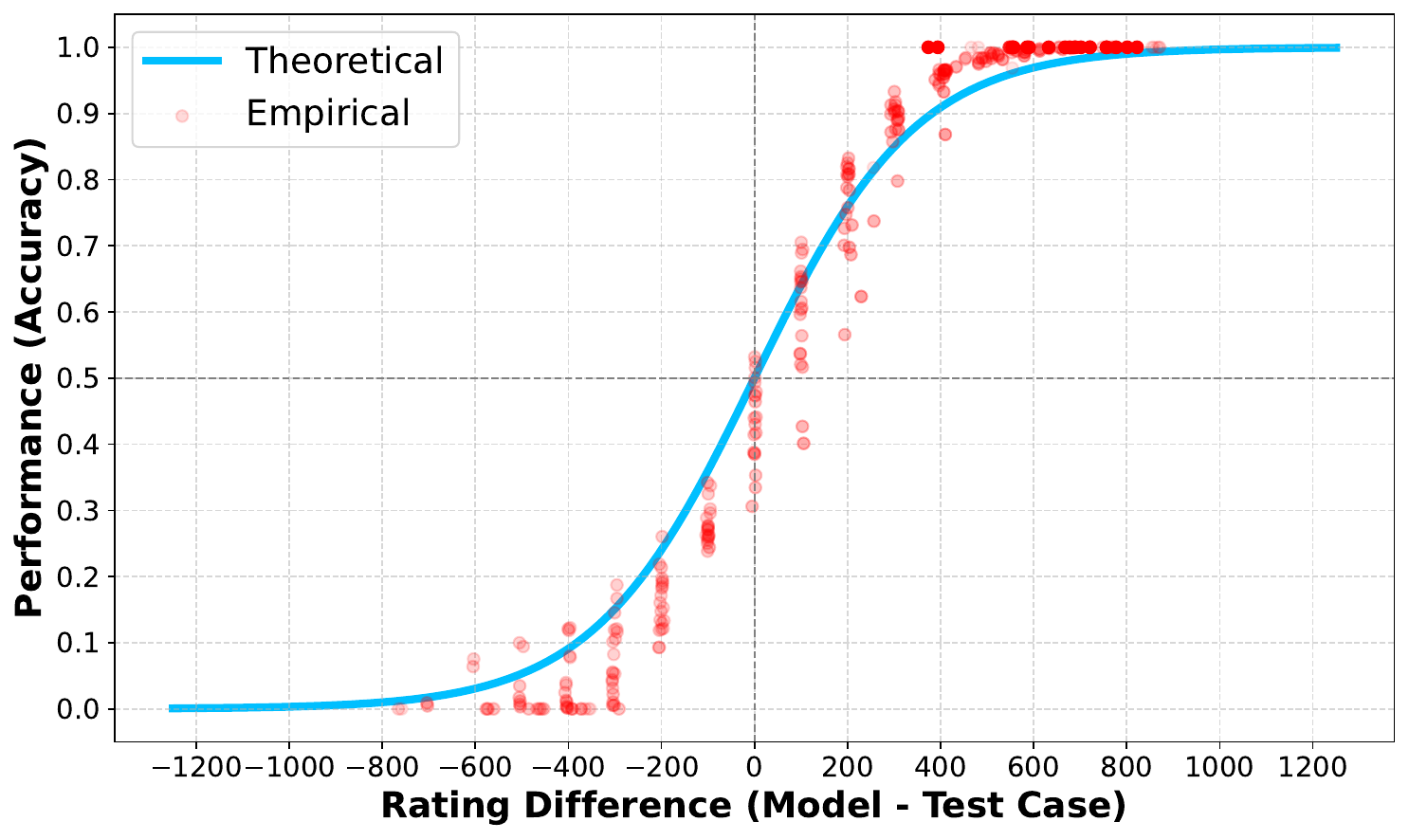}}
    \hfill
    \subfloat[Object detection: COCO~\cite{mscoco}]{\includegraphics[width=0.5\textwidth]{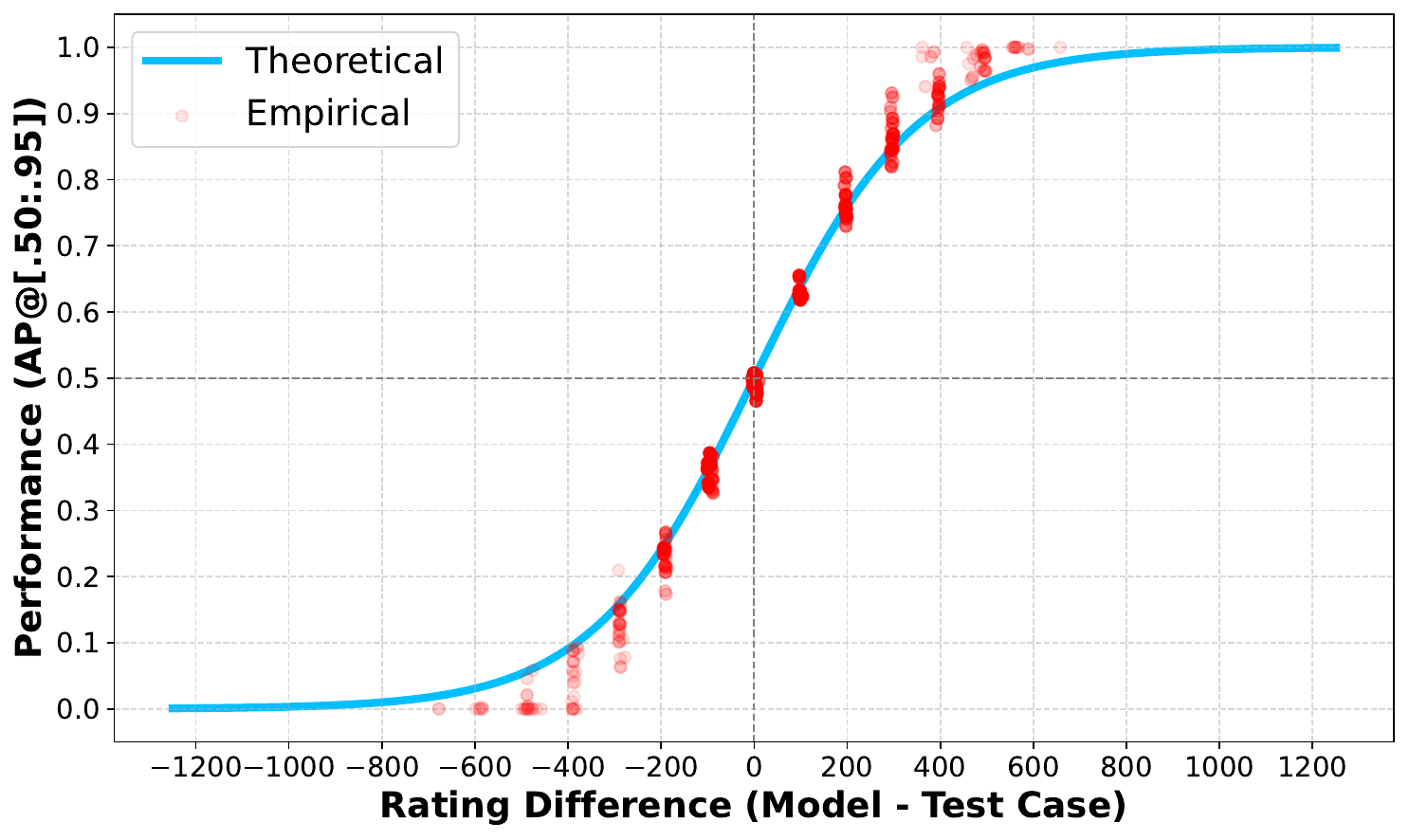}}
    \vspace{0.1pt}
    \subfloat[Question answering: MMLU~\cite{MMLU}]{\includegraphics[width=0.5\textwidth]{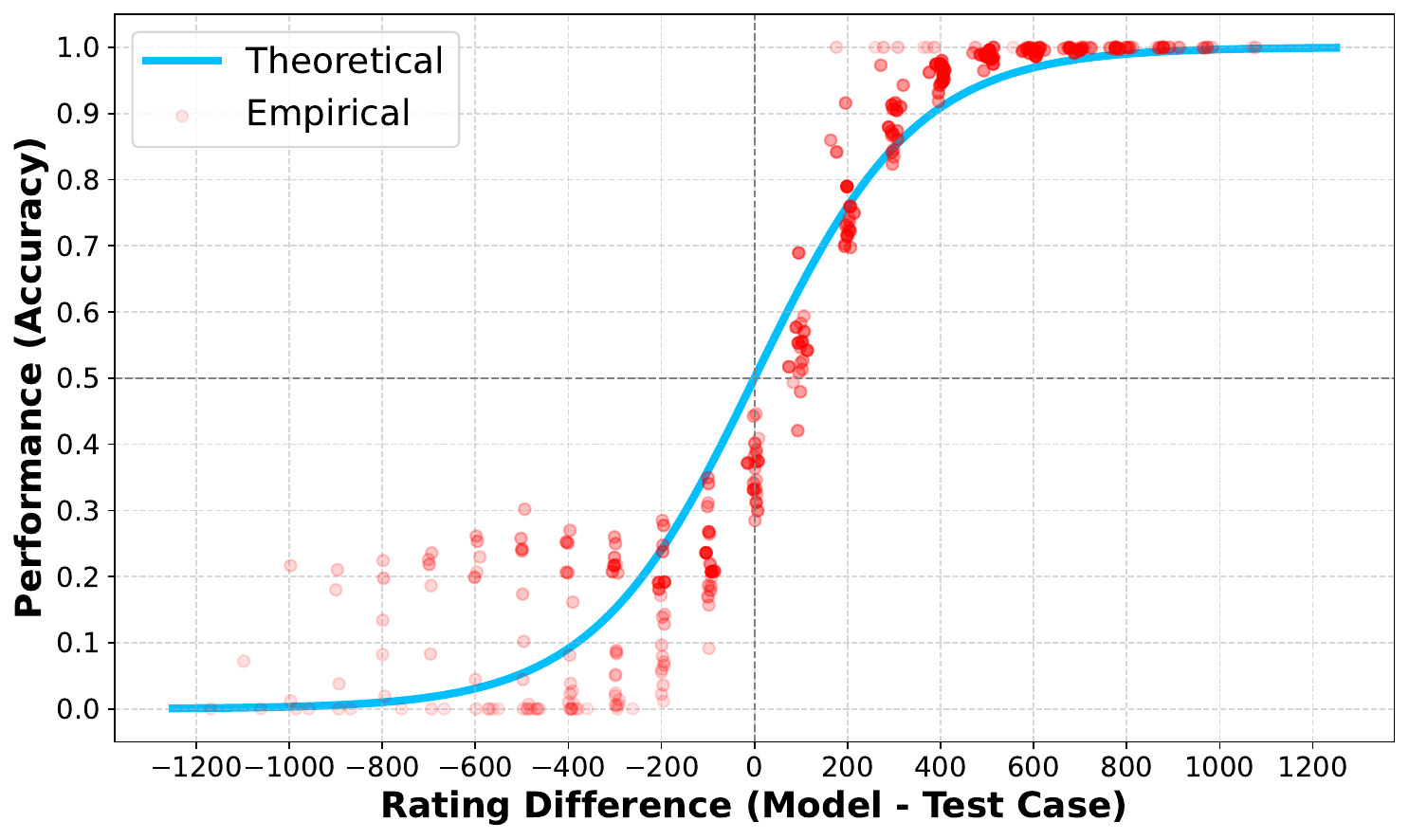}}
    \hfill
    \subfloat[Code generation: LiveCodeBench~\cite{jain2024livecodebench}]{\includegraphics[width=0.5\textwidth]{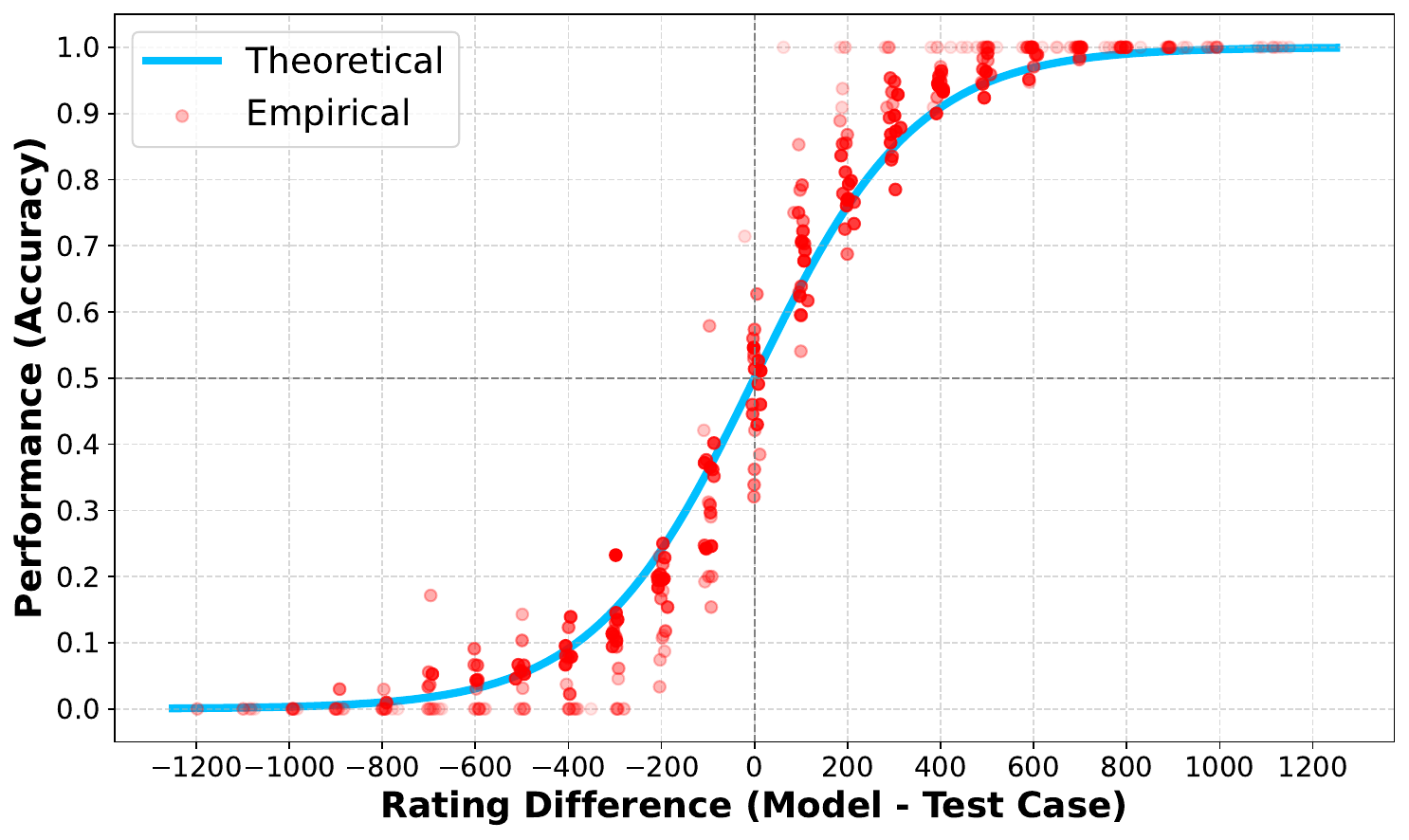}}
    \vspace{0.1pt}
    \subfloat[Motion prediction: Waymo~\cite{waymodataset}]{\includegraphics[width=0.5\textwidth]{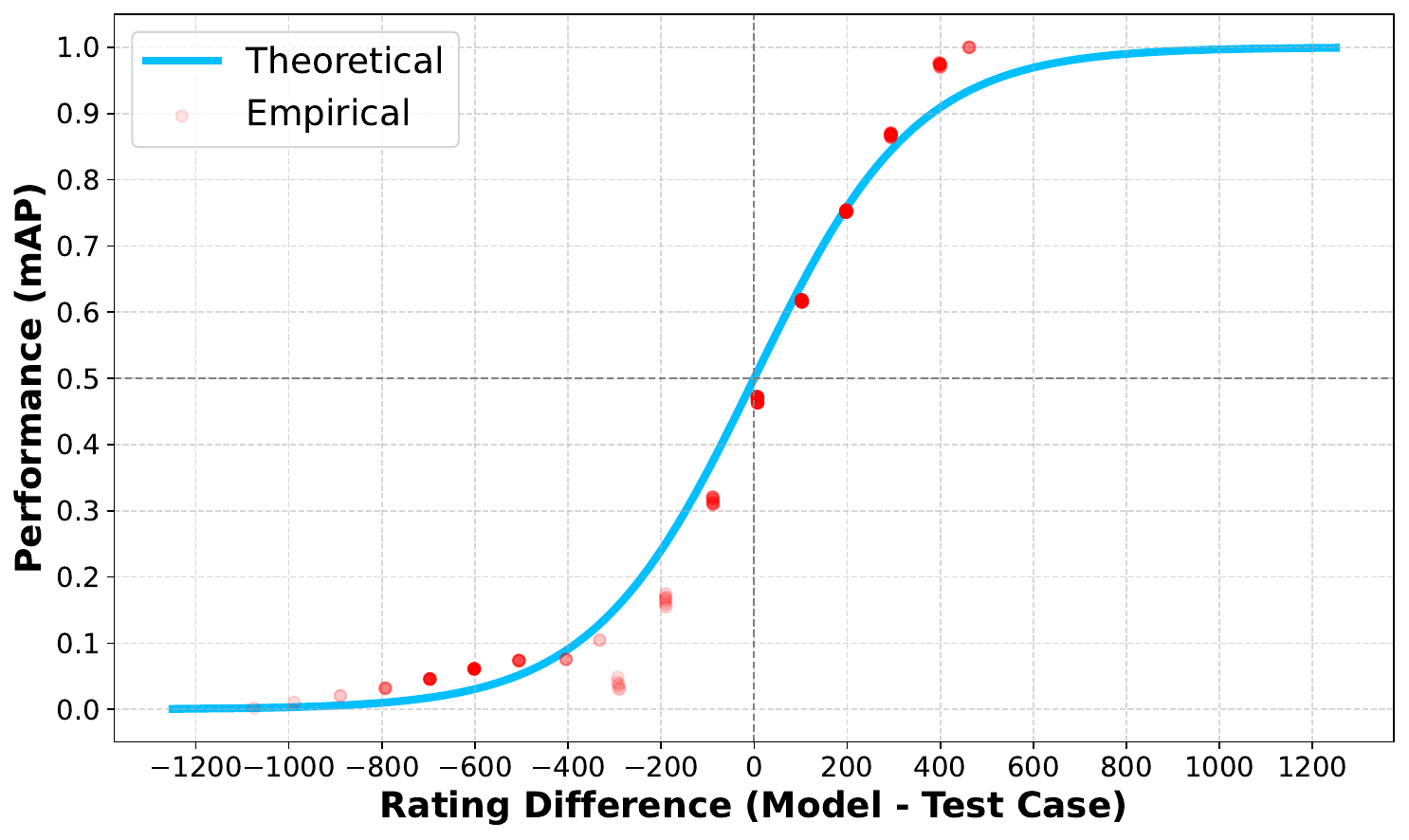}}
    \hfill
    \subfloat[Motion planning: NAVSIM~\cite{NAVSIM}]{\includegraphics[width=0.5\textwidth]{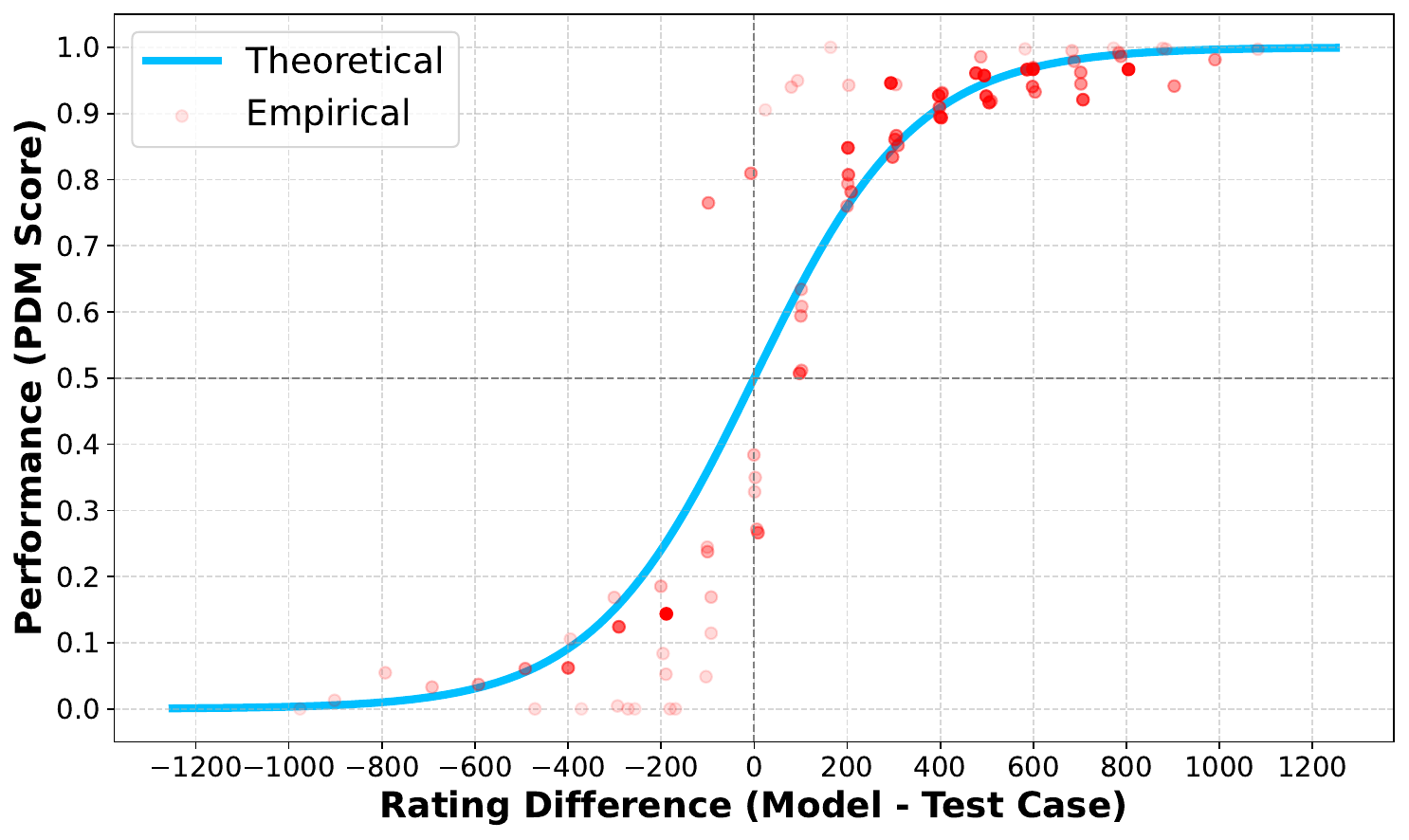}}
    
    \caption{
        Visualization of the predicted (theoretical) agent performances based on the differences between agents and test cases vs. the empirical performance obtained on each dataset. 
    }
    \label{fig:accuracy}
\end{figure}

\newpage
\subsection{Influence of match percentage on model rating stability}

\begin{figure}[ht]
    \centering
    \subfloat[Image classification: ImageNet~\cite{deng2009imagenet}]{\includegraphics[width=0.5\textwidth]{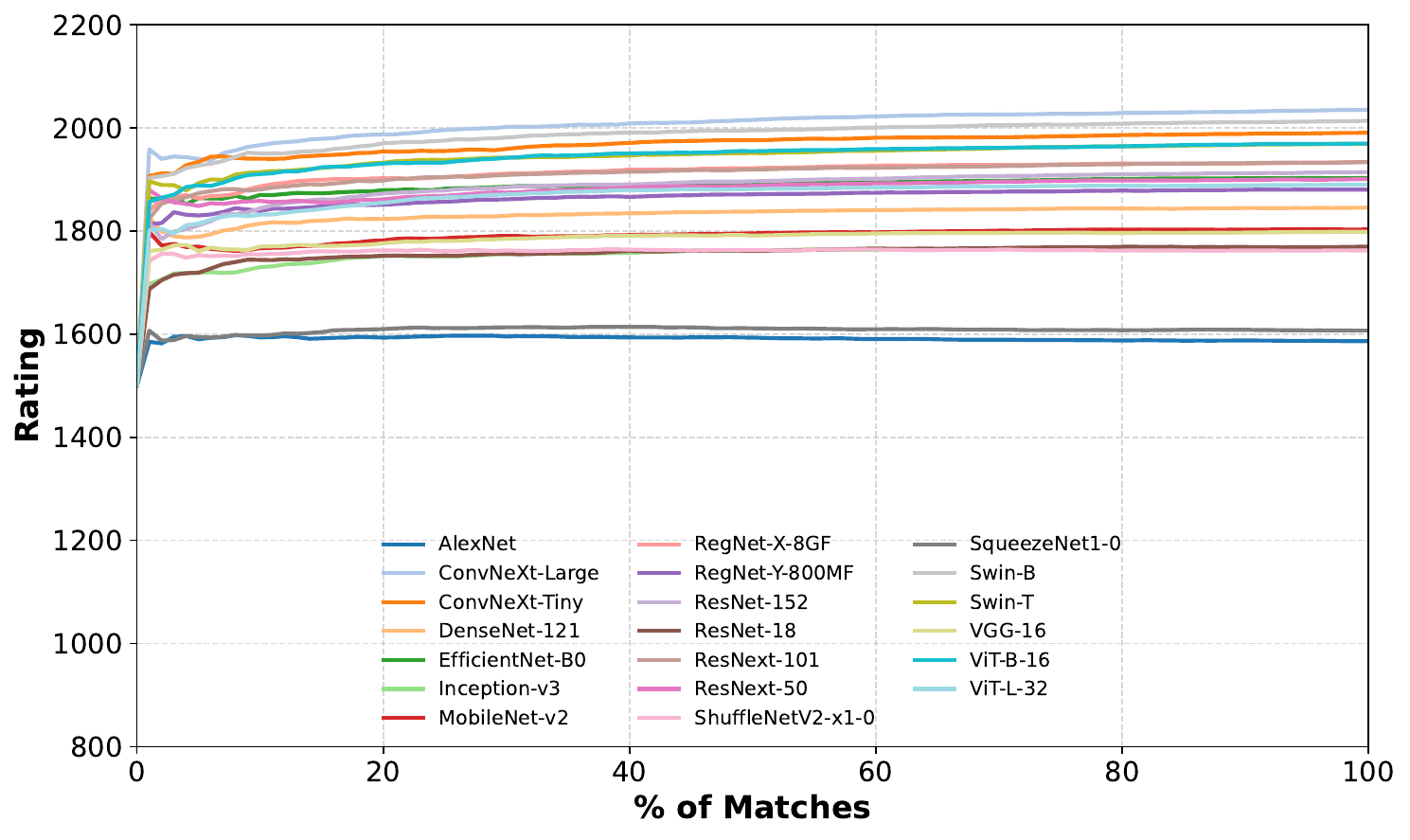}}
    \hfill
    \subfloat[Object detection: COCO~\cite{mscoco}]{\includegraphics[width=0.5\textwidth]{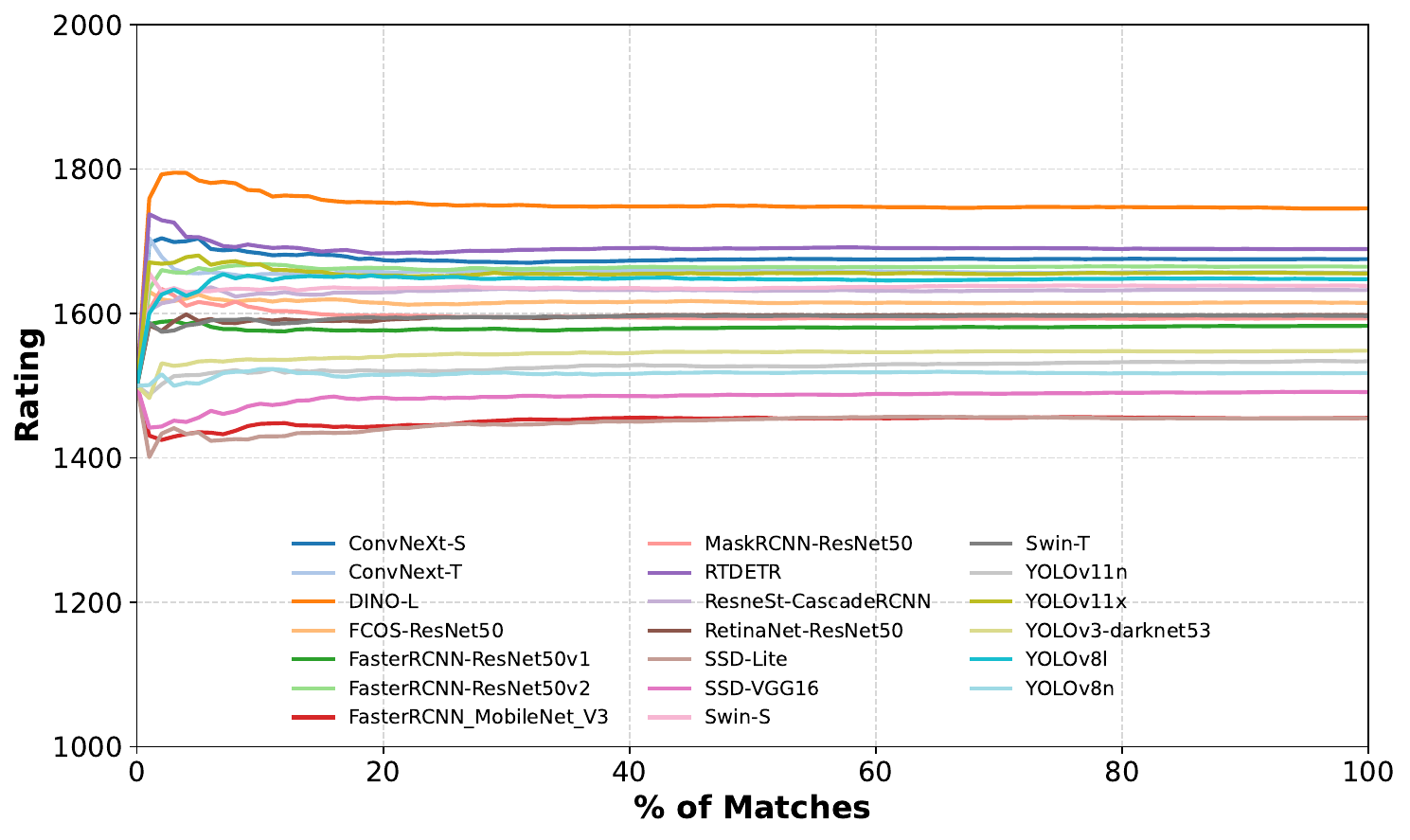}}
    \vspace{0.1pt}
    \subfloat[Question answering: MMLU~\cite{MMLU}]{\includegraphics[width=0.5\textwidth]{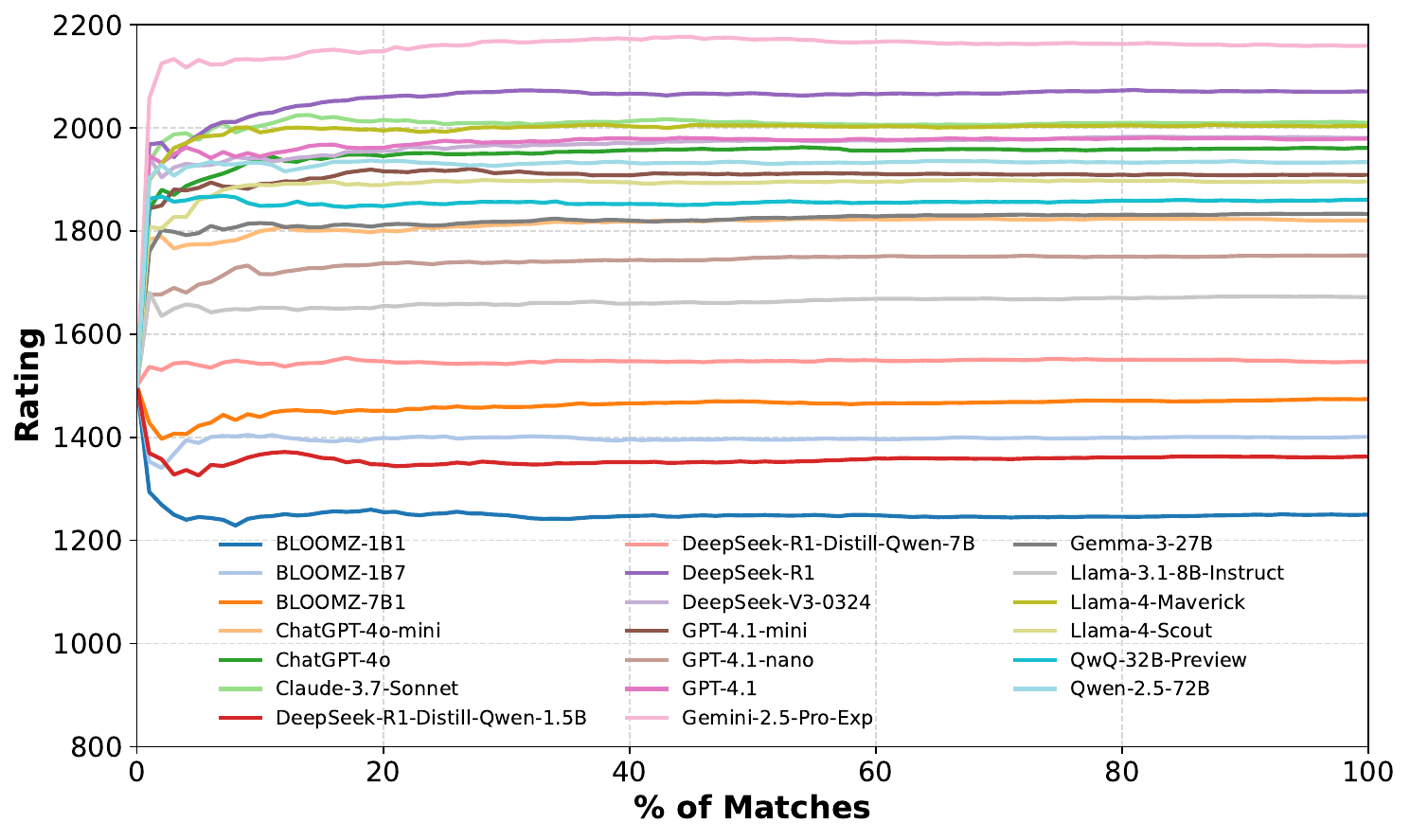}}
    \hfill
    \subfloat[Code generation: LiveCodeBench~\cite{jain2024livecodebench}]{\includegraphics[width=0.5\textwidth]{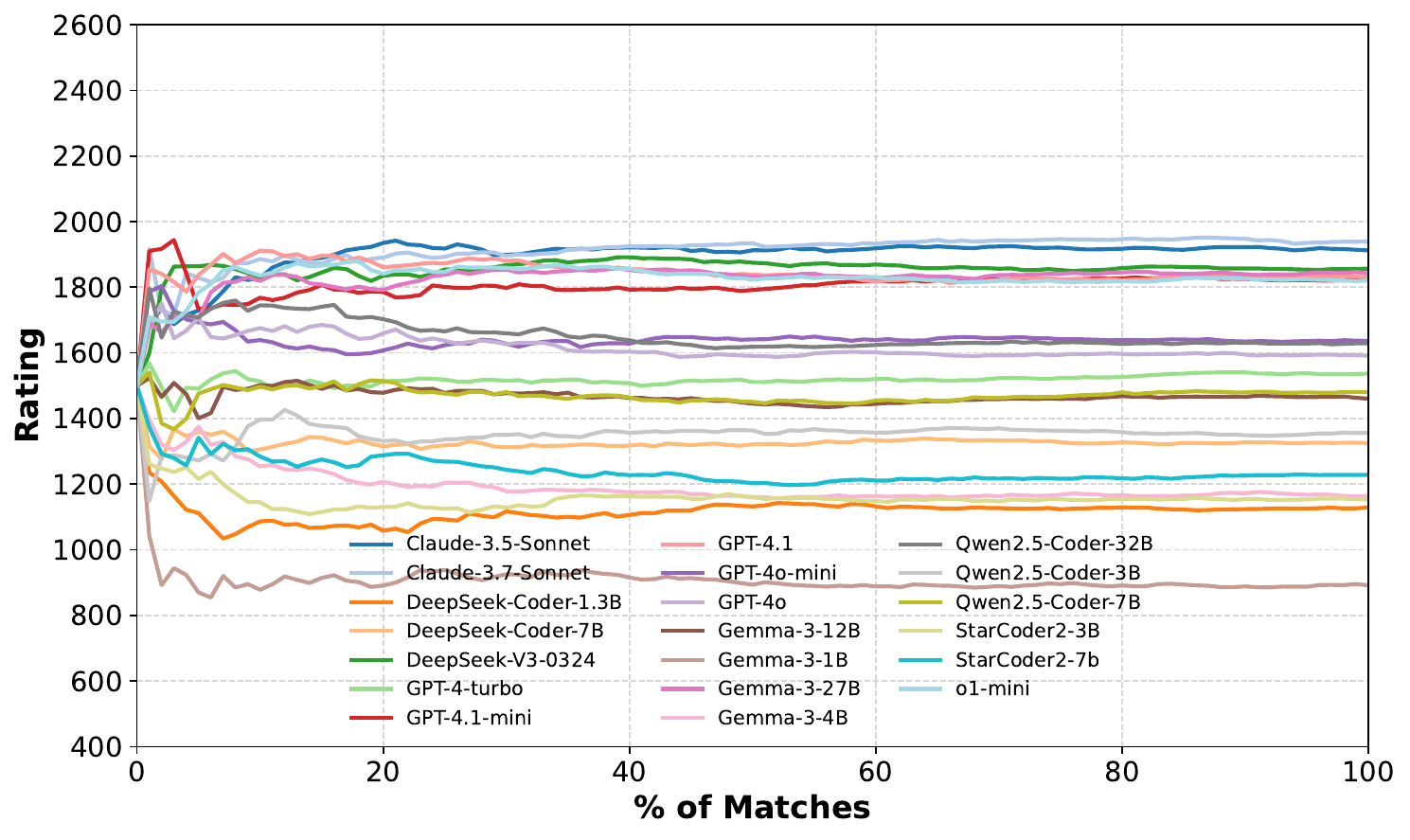}}
    \vspace{0.1pt}
    \subfloat[Motion prediction: Waymo~\cite{waymodataset}]{\includegraphics[width=0.5\textwidth]{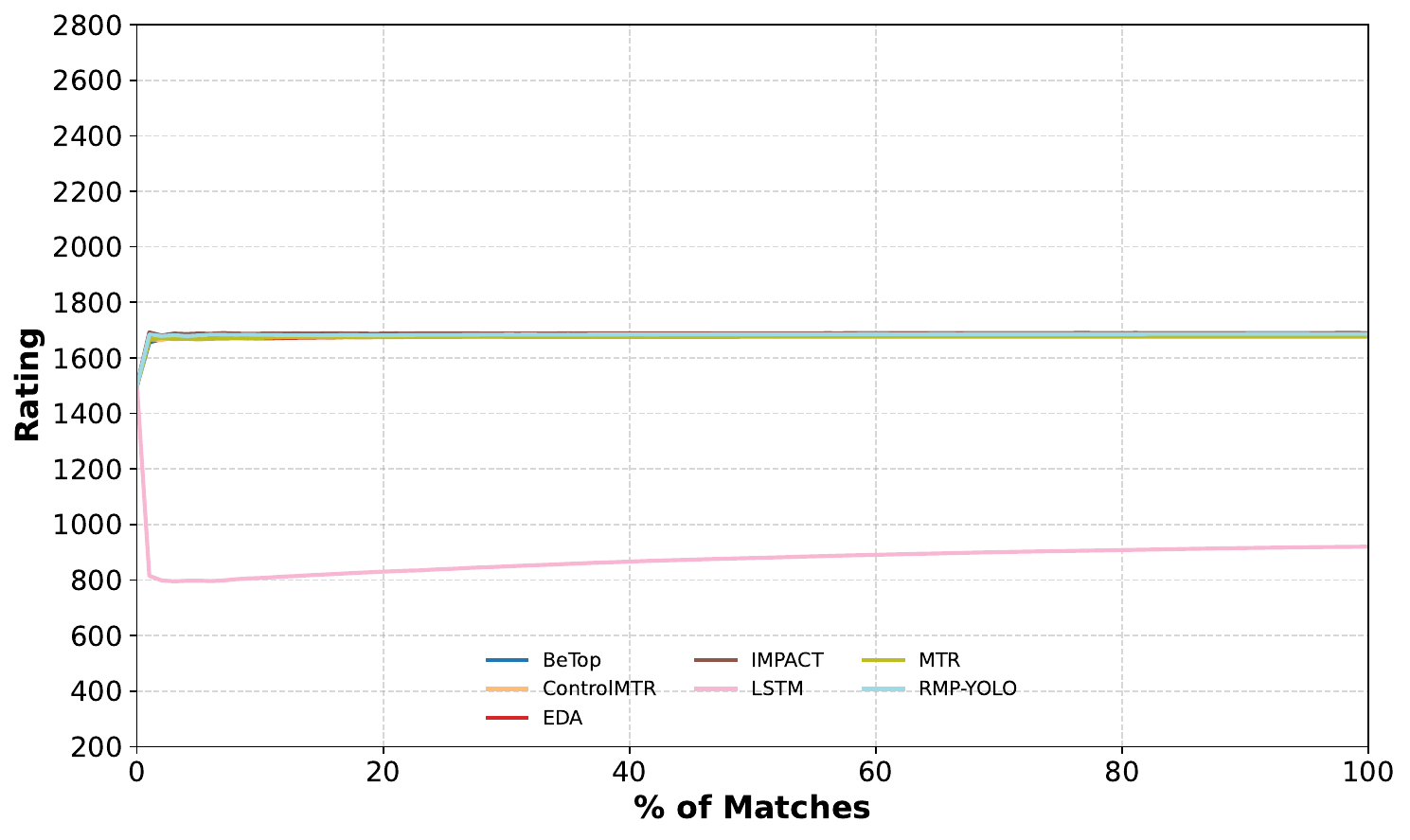}}
    \hfill
    \subfloat[Motion planning: NAVSIM~\cite{NAVSIM}]{\includegraphics[width=0.5\textwidth]{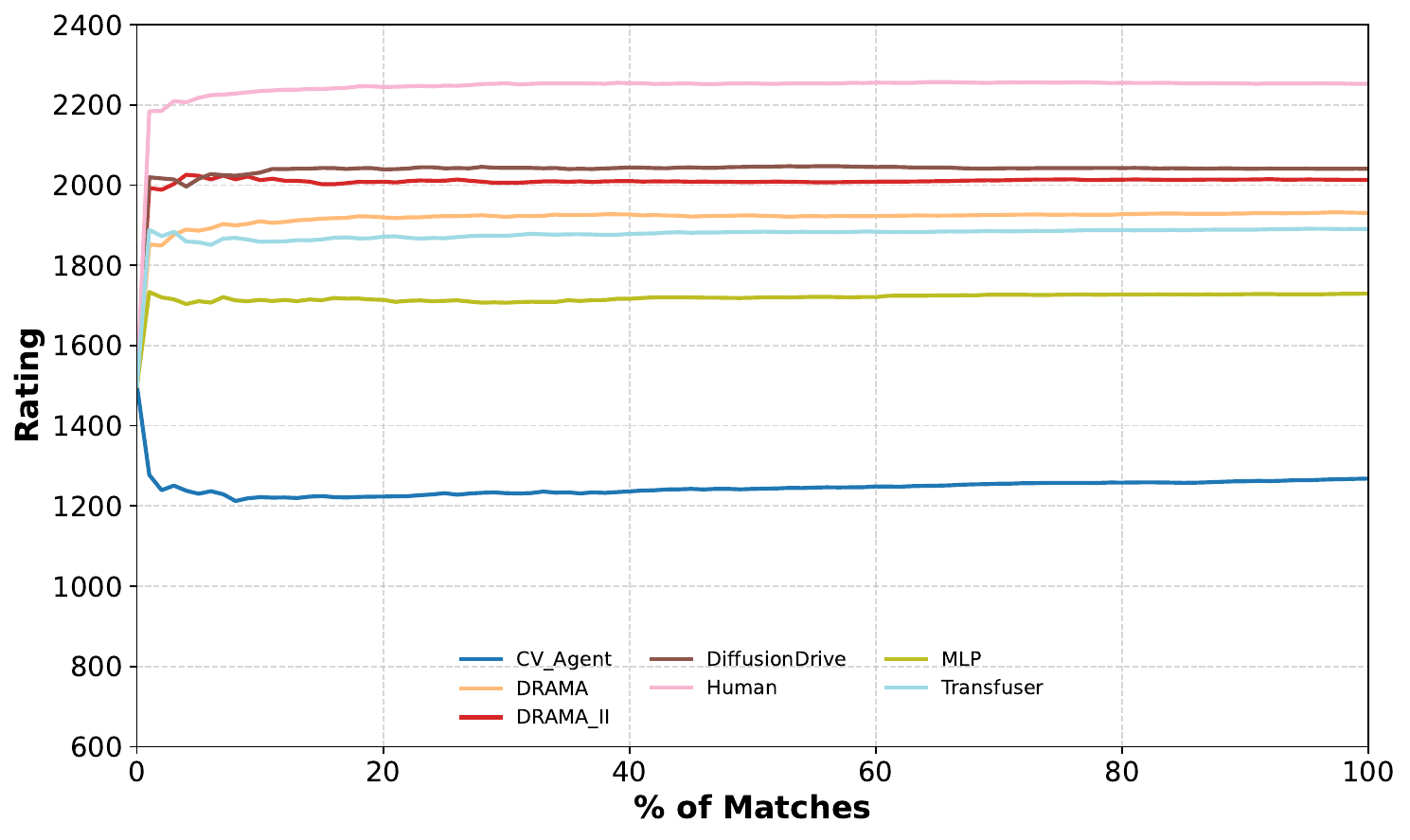}}
    
    \caption{
        Model ratings over the percentage of matches on respective datasets.
    }
    \label{fig:model_rating}
\end{figure}

\newpage
\subsection{Effect of percentage of matches on rating system accuracy and consistency}

\begin{figure}[ht]
\centering
    \subfloat[Image classification: ImageNet~\cite{deng2009imagenet}]{\includegraphics[width=0.99\textwidth]{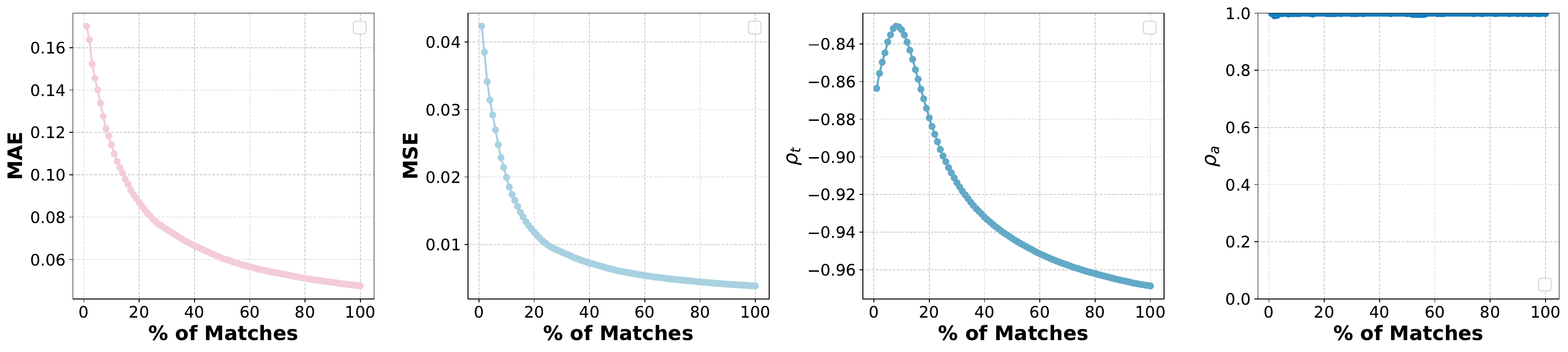}}
    \vspace{0.1pt}
    \subfloat[Object detection: COCO~\cite{mscoco}]{\includegraphics[width=0.99\textwidth]{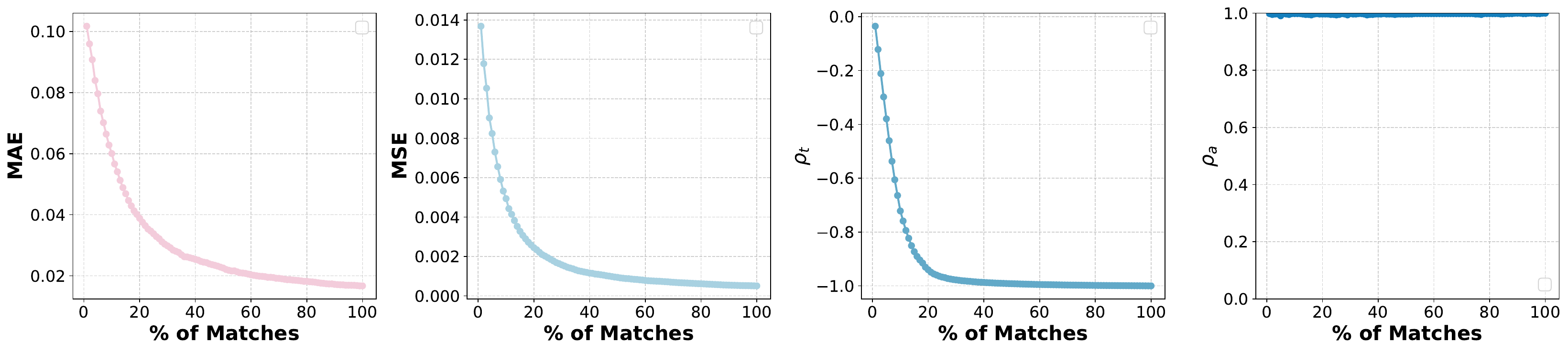}}
\caption{System prediction errors and Spearman's correlations over the percentage of matches on respective datasets (Vision).}
\label{fig:record_eval:a}
\end{figure}

\begin{figure}[ht]
\centering
    \subfloat[Question answering: MMLU~\cite{MMLU}]{\includegraphics[width=0.99\textwidth]{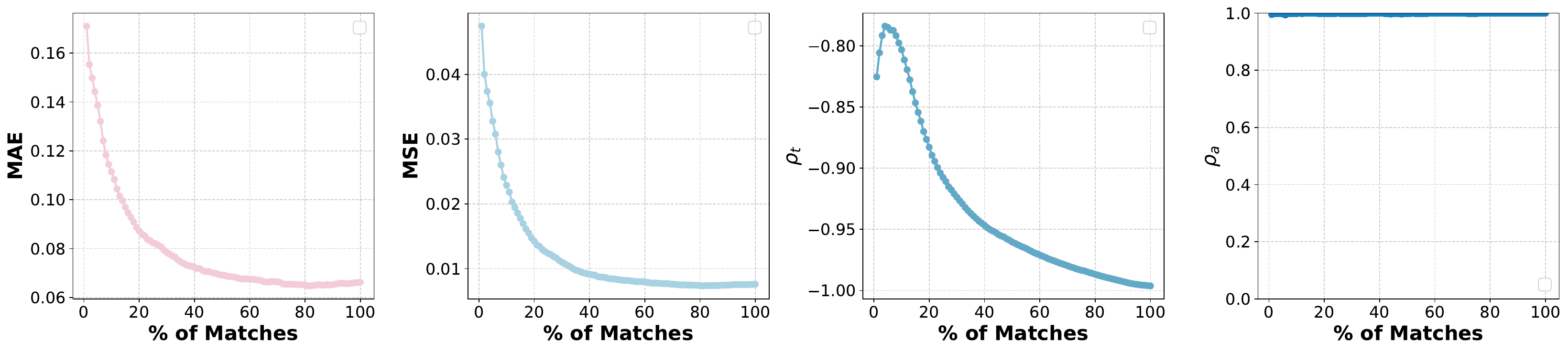}}
    \vspace{0.1pt}
    \subfloat[Code generation: LiveCodeBench~\cite{jain2024livecodebench}]{\includegraphics[width=0.99\textwidth]{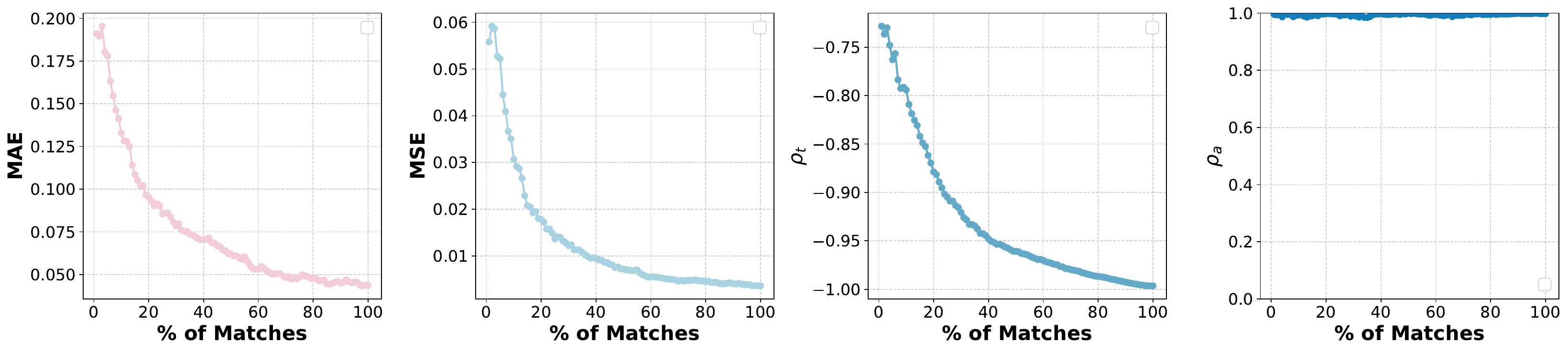}}
\caption{System prediction errors and Spearman's correlations over the percentage of matches on respective datasets (Language).}
\label{fig:record_eval:b}
\end{figure}

\newpage
\begin{figure}[ht]
\centering
    \subfloat[Motion prediction: Waymo~\cite{waymodataset}]{\includegraphics[width=0.99\textwidth]{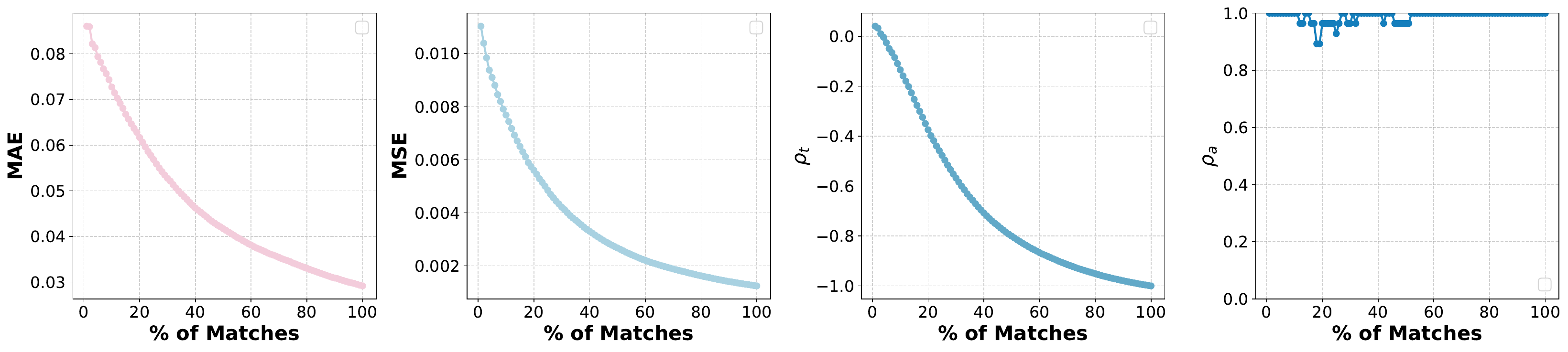}}
    \vspace{0.1pt}
    \subfloat[Motion planning: NAVSIM~\cite{NAVSIM}]{\includegraphics[width=0.99\textwidth]{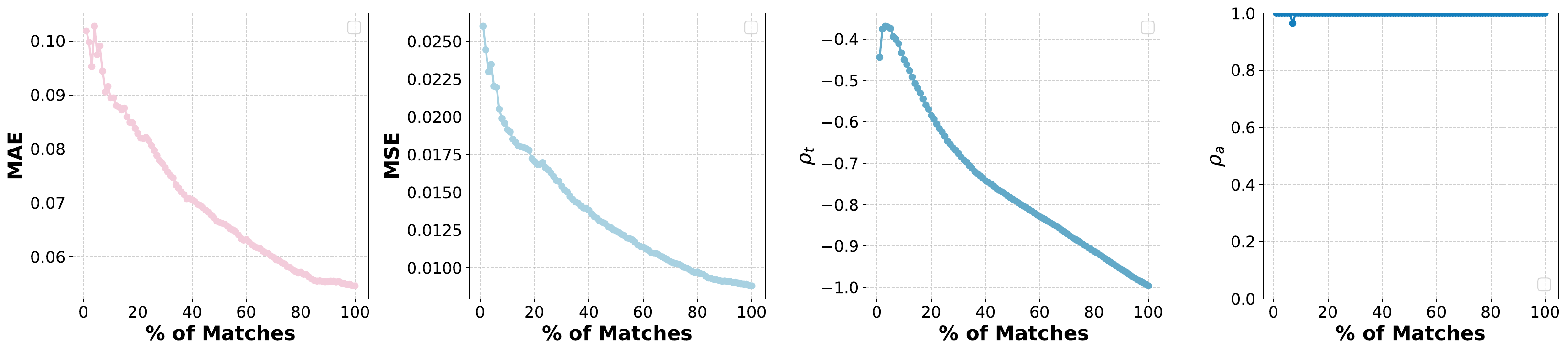}}
\caption{System prediction errors and Spearman's correlations over the percentage of matches on respective datasets (Action).}
\label{fig:record_eval:c}
\end{figure}





\newpage
\section{Detailed experimental setup}
\label{appendix:b}

\subsection{Vision - Image Classification} 

\subsubsection{Dataset} 

For the computer vision task, we selected the ImageNet~\cite{deng2009imagenet} dataset, which is one of the most widely used and challenging public benchmarks for image classification. The dataset consists of over 14 million labeled images spanning 1,000 object categories. Experiments were conducted on the validation set, which contains 50,000 distinct images, ensuring a diverse and comprehensive evaluation of model performance. 

\subsubsection{Metric}
On the ImageNet~\cite{deng2009imagenet} dataset, the standard Acc@1 metric is used:
\begin{equation}
    \text{Acc@1} = \frac{1}{N} \sum_{i=1}^{N} 1 \left( \hat{y}_i = y_i \right)
    \label{eqn:acc@1-imagenet}
\end{equation}

\subsubsection{Scoring function}
The scoring function \(f\) used on the ImageNet~\cite{deng2009imagenet} dataset is defined as:
\begin{equation}
    S := \text{Acc@1}
\end{equation}

\subsubsection{Models}
On the image classification task, we selected 20 representative image classification models and summarize their key characteristics and release years in \autoref{tab:image_classification_models}. All pretrained models were obtained from the \texttt{torchvision.models} module in PyTorch~\cite{paszke2019pytorch} and evaluated on a local desktop equipped with an Intel i9-12900K CPU, 32\,GB of RAM, and an NVIDIA RTX 3090 Ti GPU.

\begin{table}[ht]
\centering
\caption{Image classification models}
\begin{tabular}{clll}
\toprule
\textbf{\#} & \textbf{Model} & \textbf{Year} & \textbf{Source} \\
\midrule
1  & ConvNeXt-Large     \cite{liu2022convnet}           & 2022 & Pytorch \\
2  & Swin-B             \cite{liu2021swin}              & 2021 & Pytorch \\
3  & ConvNeXt-Tiny      \cite{liu2022convnet}           & 2022 & Pytorch \\
4  & ViT-B-16           \cite{dosovitskiy2020image}     & 2020 & Pytorch \\
5  & SwinT              \cite{liu2021swin}              & 2021 & Pytorch \\
6  & RegNet-X-8GF       \cite{radosavovic2020designing} & 2020 & Pytorch \\
7  & ResNext-101        \cite{xie2017aggregated}        & 2017 & Pytorch \\
8  & ResNet-152         \cite{he2016deep}               & 2016 & Pytorch \\
9  & EfficientNet-B0    \cite{tan2019efficientnet}      & 2019 & Pytorch \\
10 & ResNext-50         \cite{xie2017aggregated}        & 2017 & Pytorch \\
11 & ViT-L-32           \cite{dosovitskiy2020image}     & 2020 & Pytorch \\
12 & RegNet-Y-800MF     \cite{radosavovic2020designing} & 2020 & Pytorch \\
13 & DenseNet-121       \cite{huang2017densely}         & 2017 & Pytorch \\
14 & MobileNet-v2       \cite{sandler2018mobilenetv2}   & 2018 & Pytorch \\
15 & VGG16              \cite{simonyan2014very}         & 2014 & Pytorch \\
16 & ResNet-18          \cite{he2016deep}               & 2016 & Pytorch \\
17 & Inception-v3       \cite{szegedy2016rethinking}    & 2016 & Pytorch \\
18 & ShuffleNetV2-x1-0  \cite{ma2018shufflenet}         & 2018 & Pytorch \\
19 & SqueezeNet1-0      \cite{iandola2016squeezenet}    & 2016 & Pytorch \\
20 & AlexNet            \cite{krizhevsky2012alexnet}    & 2012 & Pytorch \\
\bottomrule
\label{tab:image_classification_models}
\end{tabular}
\end{table}

\newpage
\subsection{Vision - Object Detection} 

\subsubsection{Dataset} 

Object Detection is task started almost a decade ago. To this end, we use the establish dataset and benchmark the COCO dataset \cite{mscoco}, evaluating on the validation set, which consists of 5,000 images. 

\subsubsection{Metric}
Based on the 2017 validation split (val2017) evaluation guidelines, the metric used, AP:0.5-0.95, was calculated by averaging AP over 80 object classes AND all 10 IoU thresholds from 0.5 to 0.95 with a step size of 0.05 as shown in \autoref{eqn:coco_valap}. 

\begin{equation}
\text{AP}_{\text{COCO}} = \frac{1}{10} \sum_{k=0}^{9} \text{AP}_{\text{IoU}=0.50 + 0.05k}
\label{eqn:coco_valap}
\end{equation}

\subsubsection{Scoring function}
The scoring function \(f\) used on Waymo dataset is defined as:
\begin{equation}
    S := \text{AP}_{\text{COCO}}
\end{equation}

\subsubsection{Models}
Similar to the image classification task, we selected 20 object detection models, that vary in performance and year of development. They constitute models that have developed over the years. The models include: models with a CNN vs Transformer backbone, and vary in speed and performance. 

All pretrained models were obtained from PyTorch~\cite{paszke2019pytorch}, MMDetection~\cite{chen2019mmdetection}, and Ultralytics~\cite{jocher2020ultralytics}, and evaluated on a local desktop equipped with an Intel i9-12900K CPU, 32\,GB of RAM, and an NVIDIA RTX 3090 Ti GPU.

  
\begin{table}[ht]
\centering
\caption{Object detection models}
\begin{tabular}{clll} 
\toprule
\textbf{\#} & \textbf{Model} & \textbf{Year} & \textbf{Source} \\
\midrule
1  & DINO-L                     \cite{OD-DINO}                           & 2023 & MMDetection   \\
2  & RT-DETR                    \cite{baidu-RT-DETR}                     & 2023 & Ultralytics   \\
3  & ConvNeXt-S                 \cite{liu2022convnet}                    & 2022 & MMDetection   \\
4 & Faster R-CNN- ResNet50 -v2  \cite{OD-FasterRCNN}                     & 2015 & PyTorch   \\
5  & ConvNeXt-T                 \cite{liu2022convnet}                    & 2022 & MMDetection   \\
6  & YOLOv11x                   \cite{yolov11}                           & 2024 & Ultralytics   \\
7  & YOLOv8l                    \cite{yolov8}                            & 2023 & Ultralytics   \\
8  & Swin-S                     \cite{liu2021swin}                       & 2021 & MMDetection   \\
9 & ResNeSt                     \cite{OD-ResNeSt}                        & 2021 & MMDetection   \\
10 & FCOS                       \cite{OD-FCOS}                           & 2019 & PyTorch       \\
11 & RetinaNet                  \cite{OD-RetinaNet}                      & 2017 & PyTorch   \\
12  & Swin-T                    \cite{liu2021swin}                       & 2021 & MMDetection   \\
13 & MaskRCNN                   \cite{OD-MaskRCNN}                       & 2017 & MMDetection   \\
14 & Faster RCNN -ResNet50 - v1 \cite{OD-FasterRCNN}                     & 2015 & PyTorch   \\
15 & YOLOv3                     \cite{yolov3}                            & 2018 & MMDetection   \\
16  & YOLOv11n                  \cite{yolov11}                           & 2024 & Ultralytics   \\
17  & YOLOv8n                   \cite{yolov8}                            & 2023 & Ultralytics   \\
18 & SSD-VGG16                  \cite{OD-SSD-VGG16}                      & 2016 & PyTorch   \\ 
19 & Faster R-CNN -MobileNetv3  \cite{OD-FasterRCNN}                     & 2015 & PyTorch   \\
20 & SSDLite                    \cite{OD-SSD-VGG16}                      & 2016 & PyTorch   \\
\bottomrule 
\end{tabular}
\end{table}

\newpage
\subsection{Language - Question Answering} 

\subsubsection{Dataset} 
The MMLU (Massive Multitask Language Understanding) benchmark \cite{MMLU} is designed to evaluate models on a diverse set of challenging tasks that span 57 subjects, including mathematics, history, law, and computer science. To this end, we evaluate models on the official test split, which contains multiple-choice questions with four options each. 

\subsubsection{Metrics}
Following the original evaluation protocol, we report the Acc@1 metric, defined as the proportion of questions for which the model selects the correct answer, as shown in \autoref{eqn:acc@1-mmlu}. This metric captures the model's ability to perform zero-shot reasoning across a wide range of knowledge-intensive tasks.

\begin{equation}
    \text{Acc@1} = \frac{1}{N} \sum_{i=1}^{N} 1 \left( \hat{y}_i = y_i \right)
    \label{eqn:acc@1-mmlu}
\end{equation}

where $\hat{y}_i$ denotes the model's predicted answer for the $i$-th question, and $1(\hat{y}_i = y_i)$ is an indicator function that returns $1$ if the prediction matches the ground truth $y_i$, and $0$ otherwise.

\subsubsection{Scoring function}
The scoring function \(f\) used on Waymo dataset is defined as:
\begin{equation}
    S := \text{Acc@1}
\end{equation}

\subsubsection{Models}
For this task, we selected 20 LLMs that vary in performance and year of development.
The three BLOOMZ~\cite{muennighoff2022crosslingual} pretrained models were obtained from Huggingface~\cite{lhoest2021datasets}, and evaluated on a local desktop equipped with an Intel i9-12900K CPU, 32\,GB of RAM, and an NVIDIA RTX 3090 Ti GPU. 
The other models were evaluated using the OpenAI API~\cite{openaiAPI} and the NanoGPT API~\cite{nanogpt} online.

\begin{table}[ht]
\centering
\caption{Question answering models}
\begin{tabular}{clll}
\toprule
\textbf{\#} & \textbf{Model} & \textbf{Year} & \textbf{Source} \\ 
\midrule
1  & Gemini-2.5-Pro-Exp             \cite{google2025gemini}             & 2025 & NanoGPT API \\
2  & DeepSeek-R1                    \cite{guo2025deepseek}              & 2025 & NanoGPT API \\
3  & Claude-3.7-Sonnet              \cite{claudesonnet}                 & 2025 & NanoGPT API \\
4  & Llama-4-Maverick               \cite{meta2025llama4}               & 2025 & NanoGPT API \\
5  & DeepSeek-V3-0324               \cite{liu2024deepseek}              & 2025 & NanoGPT API \\
6  & GPT-4.1                        \cite{achiam2023gpt}                & 2025 & OpenAI API \\
7  & GPT-4o                         \cite{hurst2024gpt}                 & 2024 & OpenAI API \\
8  & Qwen2.5-72B                    \cite{yang2024qwen2}                & 2024 & NanoGPT API \\
9  & GPT-4.1-mini                   \cite{achiam2023gpt}                & 2025 & OpenAI API \\
10 & Llama-4-Scout                  \cite{meta2025llama4}               & 2025 & NanoGPT API \\
11 & QwQ-32B-Preview                \cite{qwen2024qwq32b}               & 2024 & NanoGPT API \\
12 & Gemma-3-27B                    \cite{team2025gemma}                & 2025 & NanoGPT API \\
13 & GPT-4o-mini                    \cite{hurst2024gpt}                 & 2024 & OpenAI API \\
14 & GPT-4.1-nano                   \cite{achiam2023gpt}                & 2025 & OpenAI API \\
15 & Llama-3.1-8B-Instruct          \cite{grattafiori2024llama}         & 2024 & NanoGPT API \\
16 & DeepSeek-R1-Distill-Qwen-7B    \cite{guo2025deepseek}              & 2025 & NanoGPT API \\
17 & BLOOMZ-7B1                     \cite{muennighoff2022crosslingual}  & 2023 & hf (bigscience/bloomz-7b1) \\
18 & BLOOMZ-1B7                     \cite{muennighoff2022crosslingual}  & 2023 & bigscience/bloomz-1b7 \\
19 & DeepSeek-R1-Distill-Qwen-1.5B  \cite{guo2025deepseek}              & 2025 & NanoGPT API \\
20 & BLOOMZ-1B1                     \cite{muennighoff2022crosslingual}  & 2023 & bigscience/bloomz-1b1 \\
\bottomrule
\end{tabular}
\end{table}

\newpage
\subsection{Language - Code Generation} 

\subsubsection{Dataset} 
LiveCodeBench is a recently proposed benchmark for evaluating the live code generation capabilities of large language models. To this end, we adopt the livecodebench/code\_generation\_lite dataset \cite{jain2024livecodebench}, which comprises executable, interactive coding problems designed to simulate real-world programming tasks. Evaluation is conducted on the 5th version of official test split, which contains 880 problems spanning diverse domains such as algorithms, and data structures.

\subsubsection{Metric}
Following the evaluation protocol outlined by the authors, each model is assessed based on Functional Correctness (FC), defined as the proportion \autoref{eqn:fc-codegen} of generated code completions that pass all test cases for a given problem.

\begin{equation}
    \text{FC} = \frac{1}{N} \sum_{i=1}^{N} 1 \left( \texttt{PassAll}(\hat{c}_i) \right)
    \label{eqn:fc-codegen}
\end{equation}
where $\texttt{PassAll}(\hat{c}_i)$ is an indicator function that returns $1$ if the generated code $\hat{c}_i$ passes all functional test cases for the $i$-th problem, and $0$ otherwise.

\subsubsection{Scoring function}
The scoring function \(f\) used on Waymo dataset is defined as:
\begin{equation}
    S := \text{PassAll}(\hat{c}_i)
\end{equation}

\subsubsection{Models}
For the code generation task, we selected 20 LLMs known for their strong performance in programming-related benchmarks.
Several pretrained models were obtained from Huggingface~\cite{lhoest2021datasets}, and evaluated on a local desktop equipped with an Intel i9-12900K CPU, 32\,GB of RAM, and an NVIDIA RTX 3090 Ti GPU. 
The other models were evaluated using the OpenAI API~\cite{openaiAPI} and the NanoGPT API~\cite{nanogpt} online.

\begin{table}[ht]
\centering
\caption{Code generation models}
\begin{tabular}{clll}
\toprule
\textbf{\#} & \textbf{Model} & \textbf{Year} & \textbf{Source} \\ 
\midrule
1  & Claude-3.7-Sonnet      \cite{claudesonnet}         & 2025 & NanoGPT API \\
2  & Claude-3.5-Sonnet      \cite{claudesonnet}         & 2024 & NanoGPT API \\
3  & DeepSeek-V3-0324       \cite{liu2024deepseek}      & 2025 & NanoGPT API \\
4  & Gemma-3-27B            \cite{team2025gemma}        & 2025 & NanoGPT API \\
5  & GPT-4.1                \cite{achiam2023gpt}        & 2025 & OpenAI API \\
6  & GPT-4.1-mini           \cite{achiam2023gpt}        & 2025 & OpenAI API \\
7  & o1-mini                \cite{jaech2024openai}      & 2024 & OpenAI API \\
8  & GPT-4o-mini            \cite{hurst2024gpt}         & 2024 & OpenAI API \\
9  & Qwen2.5-Coder-32B      \cite{hui2024qwen2}         & 2024 & NanoGPT API \\
10 & GPT-4o                 \cite{hurst2024gpt}         & 2024 & OpenAI API \\
11 & GPT-4-turbo            \cite{achiam2023gpt}        & 2024 & OpenAI API \\
12 & Qwen2.5-Coder-7B       \cite{hui2024qwen2}         & 2024 & hf (Qwen/Qwen2.5-Coder-7B-Instruct) \\
13 & Gemma-3-12B            \cite{team2025gemma}        & 2025 & hf (google/gemma-3-12b-it) \\
14 & Qwen2.5-Coder-3B       \cite{hui2024qwen2}         & 2024 & hf (Qwen/Qwen2.5-Coder-3B-Instruct) \\
15 & DeepSeek-Coder-7B      \cite{guo2024deepseek}      & 2024 & hf (deepseek-ai/deepseek-coder-7b-instruct) \\
16 & StarCoder2-7B          \cite{lozhkov2024starcoder} & 2024 & hf (bigcode/starcoder2-7b) \\
17 & Gemma-3-4B             \cite{team2025gemma}        & 2025 & hf (google/gemma-3-4b-it) \\
18 & StarCoder2-3B          \cite{lozhkov2024starcoder} & 2024 & hf (bigcode/starcoder2-3b) \\
19 & DeepSeek-Coder-1.3B    \cite{guo2024deepseek}      & 2024 & hf (deepseek-ai/deepseek-coder-1.3b-instruct) \\
20 & Gemma-3-1B             \cite{team2025gemma}        & 2025 & hf (google/gemma-3-1b-it) \\
\bottomrule
\end{tabular}
\end{table}


\newpage
\subsection{Action - motion prediction} 

\subsubsection{Dataset} 
For the motion prediction task, we adopt the Waymo Open Motion Dataset (WOMD)~\cite{waymodataset}, one of the most comprehensive and challenging public datasets for autonomous driving behavior prediction. WOMD is specifically designed to facilitate research on multi-agent trajectory forecasting in complex urban environments. The dataset contains a total of 486,995 training clips, 44,097 validation clips, and 44,920 testing clips. Each clip spans 8 seconds and is recorded at a sampling frequency of 10 Hz. Within each clip, 10 timesteps of historical agent states, 1 current timestep, and 80 future timesteps are provided, enabling both short-term and long-term trajectory forecasting. Evaluation is conducted on the validation split using the official Waymo evaluation API. For each selected target agent (as specified by Waymo), the model generates six candidate future trajectories along with their associated confidence scores.

\subsubsection{Metric} 

In the WOMD, there are eight predefined trajectory buckets, including straight, straight-left, straight-right, left, right, left u-turn, right u-turn, and stationary~\cite{waymodataset}. For each bucket, a predicted trajectory is classified as a false positive if it is considered a miss as defined in MR; otherwise, it is classified as a true positive. Consistent with the mAP metrics used in object detection tasks, a maximum of one true positive is assigned to the one with the highest probability, while all others are assigned a false positive. True positives and false positives are then stored by their probabilities, and a Precision / Recall (P/R) curve can be plotted for each bucket. The Average Precision (AP) is represented by the area under the P/R curve, and the mAP metric can be computed by averaging the AP across all buckets as:
\begin{equation}
    \text{mAP} = \frac{1}{N} \sum_{i=1}^{N} AP_i
\end{equation}

\subsubsection{Scoring function}
The scoring function \(f\) used on Waymo~\cite{waymodataset} dataset is defined as:
\begin{equation}
    S := \text{mAP}
\end{equation}

\subsubsection{Models}

To ensure a fair and consistent evaluation, we reproduced all listed motion prediction models using a unified hardware setup consisting of eight NVIDIA RTX 3090 GPUs. For the publicly available models, we followed their official open-source implementations closely, adapting only minor components where necessary to ensure compatibility within our evaluation framework. As ControlMTR and IMPACT is not publicly available, we contacted the authors directly and received assistance in replicating their results.
\begin{table}[ht]
\centering
\caption{Motion prediction models}
\begin{tabular}{clll}
\toprule
\textbf{\#} & \textbf{Model} & \textbf{Year} & \textbf{Source} \\
\midrule
1  & Waymo LSTM Baseline \cite{waymodataset}                               & 2021 & Proprietary \\
2  & MTR                   \cite{mtr}                                       & 2022 & https://github.com/sshaoshuai/MTR \\
3  & EDA                   \cite{lin2024eda}                                & 2023 & https://github.com/Longzhong-Lin/EDA \\
4  & ControlMTR            \cite{controlmtr}                                & 2023 & Proprietary \\
5  & RMP-YOLO              \cite{sun2024rmp}       & 2024 & https://github.com/ggosjw/RMP-YOLO \\
6  & BETOP                 \cite{liu2024betop}                              & 2024 & https://github.com/OpenDriveLab/BeTop \\
7  & IMPACT                \cite{sun2025impact} & 2025 & Proprietary \\
\bottomrule
\label{tab:motion_prediction_models}
\end{tabular}
\end{table}

\newpage
\subsection{Action - motion planning} 

\subsubsection{Dataset} 
To evaluate the motion planning performance, we adopt the NAVSIM benchmark \cite{NAVSIM}, which utilizes the OpenScene dataset \cite{contributors2023openscene} - a refined derivative of the nuPlan \cite{nuplan}. This comprehensive benchmark features 120 hours of vehicle trajectories sampled at 2Hz, providing multimodal sensor observations including: (1) synchronized 8-view high-resolution RGB image (1920×1080 pixels) and (2) fused LiDAR point clouds aggregated from five sensors. The agent's input encompasses the current observation frame along with three temporally preceding frames, thereby providing 1.5 seconds of continuous temporal context. For quantitative evaluation of the closed-loop planning performance, we employ the Predictive Driver Model Score (PDMS) provided in the NAVSIM benchmark.

\subsubsection{Metric}
The PDMS in NAVSIM v1.1 is formulated as follows:
\begin{equation}
    \text{PDMS}= \text{NC} \times \text{DAC} \times \frac{(5 \times \text{EP} + 5 \times \text{TTC} + 2 \times \text{C})}{12},
\end{equation}
where NC (no collision), DAC (driving area compliance), EP (ego progress), TTC (time-to-collision), and C (comfort) are sub-metrics as detailed in \cite{NAVSIM}.

\subsubsection{Scoring function}
The scoring function \(f\) used on NAVSIM dataset is defined as:
\begin{equation}
    S := \text{PDMS}
\end{equation}

\subsubsection{Models}
On the motion planning task, we reproduced all motion prediction models using the same hardware setup consisting of eight NVIDIA RTX 3090 GPUs. For the publicly available models, we followed their official open-source implementations closely to ensure a fair and consistent evaluation. As DRAMA II is not publicly available, we contacted the authors directly and received assistance in replicating their results.

\begin{table}[ht]
\centering
\caption{Motion planning models}
\begin{tabular}{clcl}
\toprule
\textbf{\#} & \textbf{Model} & \textbf{Year}  & \textbf{Source}\\
\midrule
1  & Human     \cite{NAVSIM}   & - & NAVSIM Ground Truth\\
2  & DiffusionDrive     \cite{diffusiondrive}   & 2025 & https://github.com/hustvl/DiffusionDrive\\
3  & DRAMA II                                   & 2025 & Proprietary\\
4  & DRAMA              \cite{yuan2024drama}    & 2024 & https://chengran-yuan.github.io/DRAMA/ \\
5  & Transfuser         \cite{Chitta2023PAMI}   & 2024 & https://github.com/autonomousvision/transfuser \\
6  & MLP                                        & 2023 & https://github.com/autonomousvision/navsim \\
7  & CV Agent                                   & 2000 & https://github.com/autonomousvision/navsim \\
\bottomrule
\label{tab:motion_planning_models}
\end{tabular}
\end{table}

\end{document}